\documentclass{article}

\usepackage{iclr2027_conference,times}
\usepackage{amsmath,amssymb,mathtools}
\usepackage{bm}
\usepackage{booktabs,multirow}
\usepackage{algorithm}
\usepackage{algorithmic}
\usepackage{graphicx}
\usepackage{etoolbox}
\usepackage{wrapfig}
\usepackage{subcaption}
\usepackage{xcolor}
\usepackage{microtype}
\usepackage{url}
\usepackage[colorlinks=true,allcolors=blue]{hyperref}

\newcommand{\missing}[1]{\textcolor{red!75!black}{\textbf{[MISSING: #1]}}}
\newcommand{\maybeinput}[2]{\IfFileExists{#1}{\input{#1}}{\missing{#2}}}
\newcommand{\pp}{\,\mathrm{pp}}

\newcommand{\mval}[2]{#1{\scriptstyle \pm #2}}
\newcommand{\bmval}[2]{\bm{#1{\scriptstyle \pm #2}}}
\usepackage{pgfplots}
\pgfplotsset{compat=1.18}

\definecolor{vlmgold}{HTML}{B8A15A}
\definecolor{vlmgreen}{HTML}{88B5A5}
\definecolor{vlmdark}{HTML}{526E68}

\title{Intervenable Concept Dynamics for Video \\ Activity Forecasting}
\author{
\textbf{Patrick Knab$^{1}$,
David Steinmann$^{2}$,
Giovanni De Felice$^{3}$,
Felix Divo$^{2}$,} \\
\textbf{Udo Schlegel$^{4}$,
Wolfgang Stammer$^{5}$,
Sascha Marton$^{1}$,
Christian Bartelt$^{1}$}
\\[0.6em]
$^{1}$Technische Universität Clausthal \quad
$^{2}$Technische Universität Darmstadt \\
$^{3}$Università della Svizzera italiana \quad
$^{4}$Ludwig-Maximilians-Universität München, MCML \\
$^{5}$Max Planck Institute for Informatics (SIC)
}
\iclrfinalcopy 
\begin{document}
\maketitle
\addtocontents{toc}{\protect\setcounter{tocdepth}{-1}}
\begin{abstract}

Concept bottleneck models route predictions through human-understandable concepts, making their intermediate states inspectable and editable. 
Extending this model family to forecasting requires modeling how concept activations evolve over time as objects, actions, and their interactions change, yet existing CBMs ground concepts only in observed inputs and do not forecast their future evolution.
We study this problem using videos of ongoing activities and forecast future activities through a named-concept state. 
We propose \textbf{TRACE}-CBM\footnote{GitHub Code: \url{https://github.com/patrick-knab/TRACE-CBM}} (Temporal Relational Activity Concept Evolution), which first refines observed concept activations with sparse spatio-temporal graphs.
It then recursively predicts future concept states, conditioning each transition on the prior activity distribution, and decodes both observed and predicted states with a shared activity classifier. 
At inference time, the model supports editing concept values, activity activations, and learned edges without parameter updates, exposing the relational spatio-temporal dependencies used for forecasting. 
Across multiple datasets, TRACE improves observed-activity accuracy over concept-based methods and stays close to matched black-box forecasters in forecast accuracy.
Intervention experiments show that concept edits correct part of the initially incorrect forecasts without flipping initially correct ones, and that activity-belief edits can steer forecasts toward a requested activity.

\end{abstract}

\section{Introduction}
\label{sec:introduction}

\textbf{Concept-based forecasting needs a semantic interface.}
Concept bottleneck models (CBMs) route predictions through human-understandable intermediate concepts \citep{koh2020concept}.
Therefore, these models expose the semantic evidence behind a prediction that supports inspection and targeted model 
\begin{figure}[h!]
\centering
\includegraphics[width=1\textwidth]{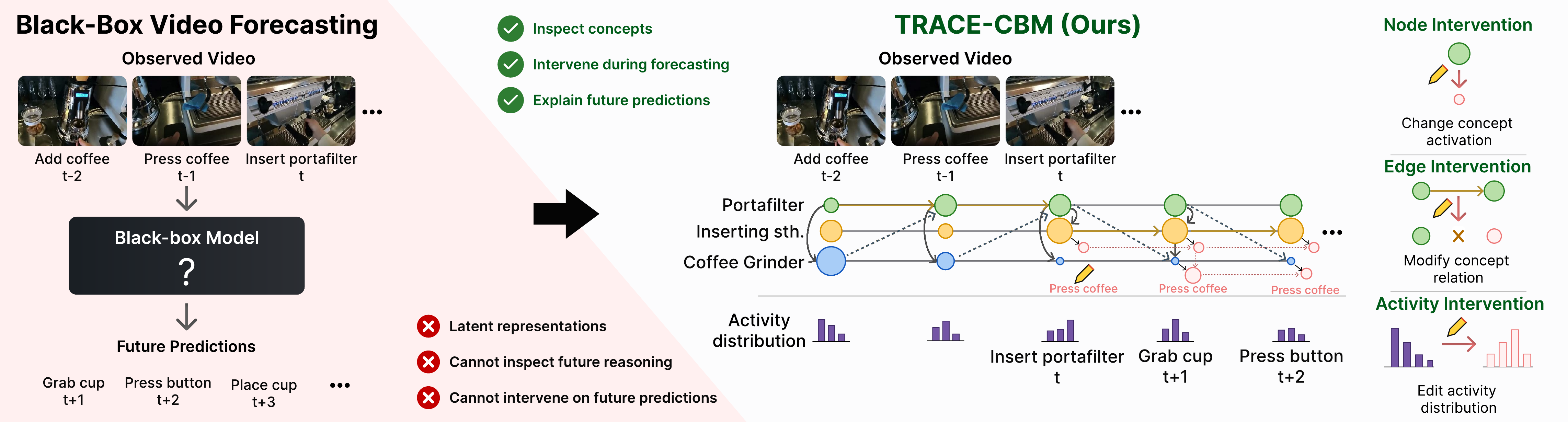}
\caption{
\textbf{Latent vs.\ concept-space forecasting.}
Conventional black-box models roll out opaque latents. TRACE-CBM rolls out editable concept states, enabling persistent edits to concepts, edges, and activity beliefs.
}
\label{fig:motivation}
\end{figure}
intervention \citep{yuksekgonul2022posthoc,yang2023language,steinmann2026objectcentric,knab2026whats}.
However, CBMs typically encode concepts as a static representation of an observed input, without modeling their evolution over time \citep{oikarinen2023labelfree,prasse2025dcbm}.
Forecasting requires modeling this evolution, as future predictions can also depend on the future semantic state.
If this evolution takes place in an opaque latent space, the semantic interface is lost, making it difficult to debug forecast errors or steer predictions at inference time \citep{felice2026position}.
An intervenable CBM for forecasting should therefore not only expose its current concept state, but preserve a meaningful and editable concept state throughout the prediction horizon.

\textbf{Videos are challenging for persistent concept forecasting.}
Activity recognition requires models not only to recognize observed video content, but also to anticipate future activities \citep{feichtenhofer2019slowfast,bertasius2021space}.
To do this, strong video forecasting models commonly propagate latent features whose future evolution cannot be directly inspected or controlled \citep{liu2022video,girdhar2021anticipative,gong2022future}.
We study whether an explicit concept interface can be maintained while semantically aligned concepts are recursively forecast into the future.
In contrast, existing video CBMs learn semantically-isolated concepts for video classification \citep{lee2025disentangled,knab2026concepts}, but do not forecast future concept states or define how interventions propagate over time.
Black-box video forecasters can predict future activities, but typically evolve opaque latent representations without a named, editable state \citep{assran2023self} (see~\autoref{fig:motivation}).
Consequently, no existing approaches provide mechanisms to provide interpretable forecasts nor have a clear mechanism to correct future concept states and thus steer future activity corrections.

\textbf{Intervenable concept-space forecasting.}
To address this limitation, we propose \textbf{TRACE} (Temporal Relational Activity Concept Evolution), an autoregressive concept bottleneck model for video activity forecasting that recursively updates named concept states, rather than opaque latent representations (see Figure~\ref{fig:motivation}).
TRACE contributes to the development of video forecasting systems that are \textit{not only accurate but also transparent and controllable}.
Our method learns spatio-temporal graphs on concept states that refine observed concept activations and represent learned within-window (spatial) and cross-window (spatio-temporal) dependencies.
We use a \textit{graph} because its \textit{explicit relation-based propagation} lets an edit to one concept propagate to other concepts, rather than treating concept edits in isolation.
The model then recursively predicts future concept states and conditions each transition on the activity distribution predicted at the previous window.
A shared concept-to-activity classifier decodes both observed and forecasted states on the same semantic space.
This design allows for model edits on concept nodes, activity beliefs, and learned edges without changing model weights.
Because the architecture exposes editable concept states during forecasting, we evaluate whether these edits can correct or steer future activity predictions.
Our main contributions are as follows:

\begin{itemize}

\item We introduce TRACE, an autoregressive CBM for video forecasting that predicts human-readable concept states with sparse spatio-temporal graphs and activity-conditioned transitions (Section~\ref{sec:method}).

\item Because forecast states remain in the same concept space as observed states, interventions on concepts, activity beliefs, and learned relations can be \textit{traced} and propagated over time, extending CBM interventions to future states without retraining (Section~\ref{sec:method}).

\item We show experimentally that graph-structured dynamics preserve localized, inspectable concept contributions, and quantify how edits to concepts, activity beliefs, and relations propagate through the forecasting process (Section~\ref{sec:evaluation}).

\end{itemize}

\section{Related work}
\label{sec:related}

\textbf{Concept bottleneck models.}
Concept bottleneck models factor prediction into an input-to-concept encoder and a concept-to-target decoder, such that a named concept vector controls the task decision \citep{koh2020concept,zarlenga2022concept,yuksekgonul2022posthoc, knab2026whats}.
Subsequent work reduces manual concept supervision through label-free concepts \citep{oikarinen2023labelfree}, improves concept construction using language guidance \citep{yang2023language}, and explores data-efficient \citep{prasse2025dcbm} and object-centric representations \citep{steinmann2026objectcentric}.
Beyond interpretability, concepts provide an interface for a targeted model intervention.
For example, \citet{shin2023intervention} study how correction gains depend on the concept selection and the intervention budget, while \citet{zarlenga2023intcem} explicitly train concept embedding models for sequences of human corrections.
However, these interventions primarily modify concepts underlying a current prediction. 
Standard CBM factorization does not specify how an edit should propagate through recursively predicted future concept states.
Concept representations can also retain target-relevant information beyond their intended semantics \citep{havasi2022addressing}, motivating leakage-aware construction and evaluation.

\textbf{Temporal modeling and forecasting.}
Modern video recognition and anticipation models capture temporal structure through convolutional or attention-based representations, including SlowFast \citep{feichtenhofer2019slowfast} and TimeSformer \citep{bertasius2021space}.
Anticipative Video Transformer \citep{girdhar2021anticipative} and FUTR \citep{gong2022future} extend such representations toward future activity prediction from partial observations.
While effective for temporal prediction, their propagated states do not provide a named semantic interface for inspection or intervention \citep{felice2026position}.
In contrast, recent video CBMs have begun to incorporate temporal concept representations.
DANCE \citep{lee2025disentangled} separates explanatory factors, while \citet{knab2026concepts} preserve temporally ordered concept representations with MoTIF.
These methods support concept-based temporal video understanding but do not forecast future concept states or enable interventions that persist throughout the forecast.
Our TRACE architecture instead uses named concepts as the forecasting state, allowing it to predict future activities and propagate the effects of edits over time.

\textbf{Structured concept dynamics.}
Relations among concepts provide another direction beyond the standard independent-concept representation.
Graph CBMs explicitly model dependencies between concept variables \citep{xu2026graph}, while ReCBM uses uncertainty-gated semantic relations to refine unreliable concept states \citep{sui2026recbmuncertaintygatedrelationalreasoning}.
Causally reliable CBMs and Causal Concept Graph Models further investigate concept relations aligned with real-world causal mechanisms \citep{de2026causally,dominici2025causal}.
Across these approaches, relational structure primarily supports concept reasoning or refinement at a given prediction point.
In contrast, TRACE-CBM uses sparse spatio-temporal graphs as an inspectable predictive mechanism: it both refines observed concepts and propagates named concept states through the forecast horizon.
Consequently, edits to concept nodes, relations, or activity beliefs can persist throughout the forecast trajectory.
\section{Method}
\label{sec:method}

\begin{figure}[t]
    \centering
    \includegraphics[width=\textwidth]{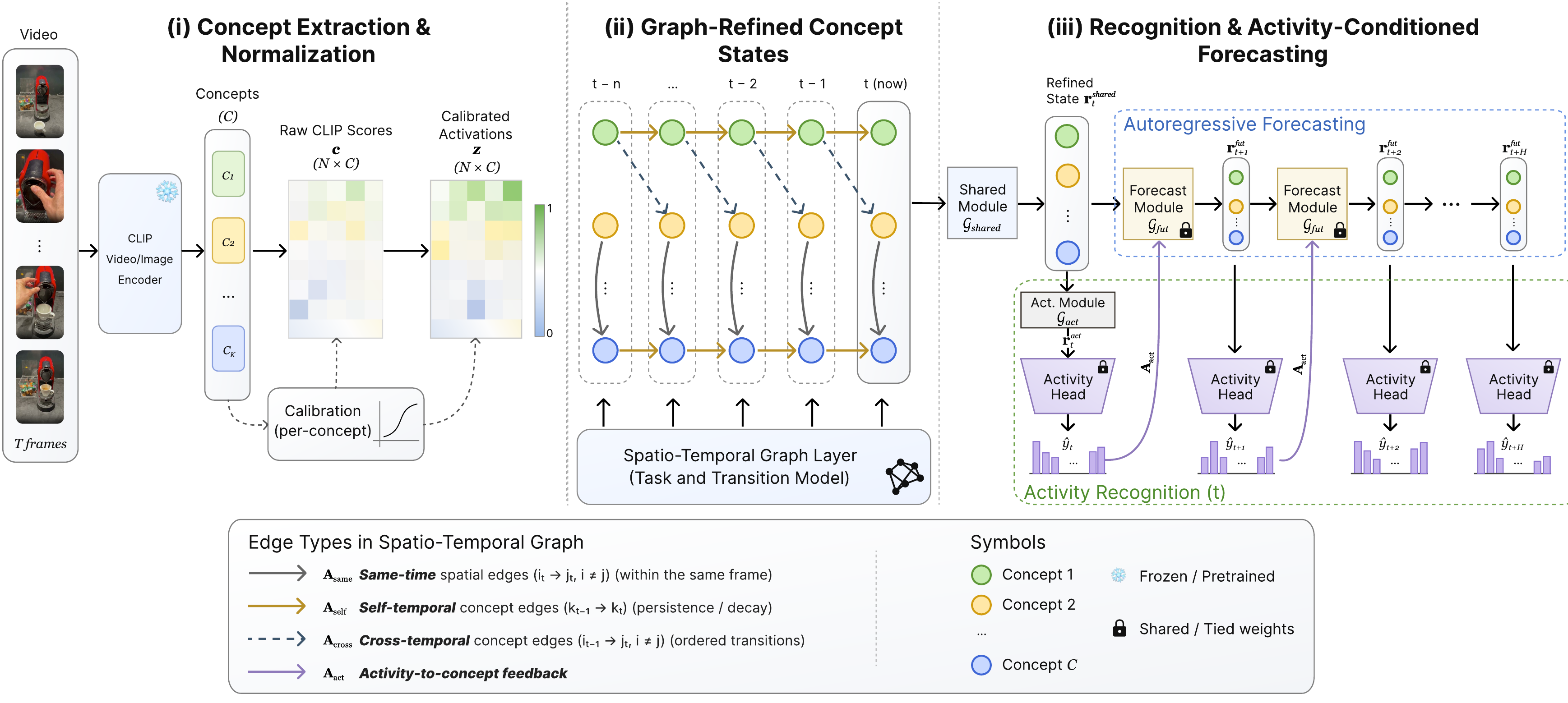}
    \caption{
    \textbf{Overview of the TRACE  pipeline.}
    Normalized concept states are processed by shared-refinement, recognition, and forecast graph modules and decoded by a shared activity head.
    Activity-to-concept feedback refines the observed state and conditions recursive concept forecasting, making concept, edge, and activity-belief interventions explicit model operations.
    }
    \label{fig:architecture}
\end{figure}

While concept bottleneck models utilize named intermediate concepts, their standard formulation does not specify how these concepts change during forecasting or how interventions (edits) to the current model state should propagate to future concept states.
\textbf{Research question:}
\emph{Can future predictions retain a named, editable state by forecasting directly in concept space while modeling interactions between concepts over time?}
\textbf{TRACE} (Figure~\ref{fig:architecture}) refines observed concepts through normalization, sparse spatio-temporal graphs and then recursively forecasts future, editable concept states, conditioned on preceding activity beliefs.
A shared linear predictor (activity head) maps observed and forecast states to activity predictions, enabling an interpretable activity forecasting pipeline by design.

\subsection{Concept extraction and normalization}
\label{sec:problem}

Let a video $x$ consist of several frames partitioned into $N$ equal, non-overlapping windows, each with one of $K$ activity labels: $y_t \in \{1, \dots, K \}$.
At prediction time \(T\), the model gets the observed sequence \(x_{1:T}\) and predicts the current activity \(y_T\) and future activities \(y_{T+h}\) for \(h\in\{1,\ldots,H\}\), where \(T+H\le N\).

\textbf{Concept extraction.}
First, we represent each video window through human-understandable concept evidence.
We define a vocabulary $\mathcal{C}$ of $C=|\mathcal{C}|$ concepts as short natural-language phrases~\citep{yang2023language} and embed these concepts and the video windows into a shared vision–language space using CLIP-like models~\citep{clip,bolya2025perception}.
For window embedding $\mathbf{v}_t$ at time $t$ and concept-text embedding $\mathbf{e}_j$ of concept $j$, the raw concept score is their cosine similarity,
$c_{t,j}=\mathbf{v}_t^\top\mathbf{e}_j/(\lVert\mathbf{v}_t\rVert\lVert\mathbf{e}_j\rVert)$.
The offline vocabulary-construction prompt and verification procedure are provided in Algorithm~\ref{alg:vlm-concept-generation} and Appendix~\ref{app:vlm-concepts}, with details on video embeddings in Appendix~\ref{app:dataset-baseline-details}.
After training-split standardization, $\mathbf{c}_t \in \mathbb{R}^C$ are the concept scores for the $t$-th window.
We treat these named similarity vectors as observations of human-readable concepts rather than anonymous latent vectors.
To cleanly batch video sequences of different length, we distinguish observed windows from padding with a validity indicator $m_t\in\{0,1\}$, where $m_t=1$ for an observed window and $m_t=0$ for padding.
Padded windows do not affect subsequent computations or losses.

\textbf{Concept-score normalization.} 
Following VLM-based CBMs \citep{yang2023language, oikarinen2023labelfree}, we use image–text similarity as concept evidence~\citep{oikarinen2023labelfree}, but map the raw scores to bounded concept states \citep{koh2020concept}. 
Since similarity-score distributions can differ across concepts, we learn a separate threshold $\boldsymbol\tau$ and a positive scaling factor $\mathbf{s}$ for each concept.
For each valid window $t$, we compute
\begin{equation}
    \widetilde{\mathbf{z}}_t=
    \sigma\!\left(
        (\mathbf{c}_t-\boldsymbol\tau)\odot\mathbf{s}
    \right)\in(0,1)^C,
    \qquad
    \mathbf{z}_t=
    \begin{cases}
        \widetilde{\mathbf{z}}_t, & m_t=1,\\
        \mathbf{0}, & m_t=0.
    \end{cases}
    \label{eq:calibration}
\end{equation}
where, $\sigma$ denotes the sigmoid function and $\odot$ denotes elementwise multiplication.

\subsection{Graph-refined concept states}
\label{sec:graph-refinement}

TRACE performs message-passing computations directly on spatio-temporal graphs of named concept states, allowing each concept to integrate information from related concepts and from its own states over time \citep{yan2018spatial}.
Edges in the graph are learned sparse dependencies within and across windows without spatial image structure.
We first introduce the graph modules used in TRACE (Figure~\ref{fig:architecture}) before defining the common graph architecture underlying them.
Following the pipeline from left to right, the graph in the center illustrates an example of the refined concept states and relations represented by a graph module.

\textbf{Graph modules.} 
The pipeline comprises three modules: a shared-refinement module \(\mathcal{G}_{\mathrm{shared}}\) that refines the sequence of concept states into a common representation $\mathbf{r}$, an activity module \(\mathcal{G}_{\mathrm{act}}\) that specializes this representation for current-activity prediction, and a forecasting module \(\mathcal{G}_{\mathrm{fut}}\) that recursively predicts futurestates.
Each of their observed states are
\begin{equation}
    \mathbf{r}^{\mathrm{shared}}_{1:T}
    =\mathcal{G}_{\mathrm{shared}}(\mathbf{z}_{1:T}),
    \qquad
    \mathbf{r}^{\mathrm{act}}_{1:T}
    = \mathcal{G}_{\mathrm{act}}(\mathbf{r}^{\mathrm{shared}}_{1:T}),
    \qquad
    \mathbf{r}^{\mathrm{fut}}_{1:t+1}
    =\mathcal{G}_{\mathrm{fut}}(\mathbf{r}^{fut}_{1:t}). 
    \label{eq:graph-stacks}
\end{equation}

The activity module $\mathcal{G}_{\mathrm{act}}$ updates the current-activity state based on $\mathbf{r}^{\mathrm{shared}}_{1:T}$ at $h=0$, whereas $\mathcal{G}_{\mathrm{fut}}$ recursively predicts future concept states, starting from $\mathbf{r}^{fut}_{1:T} = \mathbf{r}^{\mathrm{shared}}_{1:T}$. 
For $h\geq1$, the activity head predicts the future activity labels $\hat{\mathbf{y}}_t$ from the $\mathcal{G}_{\mathrm{fut}}$-updated forecast state without reapplying $\mathcal{G}_{\mathrm{act}}$.

\textbf{Spatio-temporal concept graph module.} A node $(t,j)$ in the graph represents concept $j$ at window $t$, with initial state $z_{t,j}$. 
The graph modules perform message passing over the sequence of concept states and update their representations. 
Thereby, $\mathcal{G}^{(l)}$ denotes one graph message-passing layer at depth $l$, and throughout this paper, we use a single layer ($l=1$) for all modules, but additional layers can be added.
Each module operates on the full sequence and applies the same local relation patterns at every window:
$\mathbf{r}^{(0)}_{1:t}$ is full the module-specific input state (Eq.~\ref{eq:graph-stacks}), and 
$\mathbf{r}^{(l)}_{1:t}=\mathcal{G}^{(l)}\!\left(\mathbf{r}^{(l-1)}_{1:t}\right)$.
Each module contains three relation families (Figure~\ref{fig:architecture}): (i) \emph{same-time} relations,
$\mathbf{A}^{(l)}_{\mathrm{same}}\in\mathbb{R}^{C\times C}$,
between distinct concepts at the same time window; (ii) \emph{self-temporal} relations,
$\mathbf{a}^{(l)}_{\mathrm{self}}\in\mathbb{R}^{C}$,
carrying each concept from $t-1$ to itself at $t$; and (iii) \emph{cross-temporal} relations,
$\mathbf{A}^{(l)}_{\mathrm{cross}}\in\mathbb{R}^{C\times C}$,
carrying one concept at $t-1$ to a different concept at $t$.
Temporal edges connect only adjacent windows, so longer-range dependencies arise by composing these local updates across time and message-passing depth.
Because concept states are bounded between $0$ and $1$, the modules applies additive updates with logit transformation $\phi(\mathbf{u})=\log(\mathbf{u}/(1-\mathbf{u}))$ and maps the result back to $(0,1)^C$ with a sigmoid. 
For target window $t$, a spatio-temporal concept graph module $l$ computes
\begin{equation}
    \mathbf{r}^{(l)}_t
    =
    \sigma\!\left(
        \phi\!\left(\mathbf{r}^{(l-1)}_t\right)
        +
        \boldsymbol\lambda_{\mathcal{G}}^{(l)}\odot
        \left[
            \mathbf{r}^{(l-1)}_t\mathbf{A}^{(l)}_{\mathrm{same}}
            +
            \mathbf{r}^{(l-1)}_{t-1}
                \odot\mathbf{a}^{(l)}_{\mathrm{self}}
            +
            \mathbf{r}^{(l-1)}_{t-1}
                \mathbf{A}^{(l)}_{\mathrm{cross}}
        \right]
    \right),
    \label{eq:graph-layer}
\end{equation}

where $\boldsymbol\lambda_{\mathcal{G}}^{(l)}$ is a learned per-concept update gate. 
The first term preserves an identity path for every concept coordinate, whereas the remaining terms aggregate relational messages. 
We apply top-$q$ pruning to the additive messages, and leave the identity path unpruned. 
For same-time and cross-temporal relations, each target retains its strongest incoming sources under the sparsification parameters $q_{\mathrm{same}}$ and $q_{\mathrm{cross}}$, respectively, and for self-temporal relations, $q_{\mathrm{self}}$ retains the strongest same-concept channels.
The sparsity budgets are ramped up during training rather than imposed from the start: pruned relations contribute zero, while the weights of retained relations remain differentiable.

\subsection{Recognition and activity-conditioned forecasting}
\label{sec:recognition-forecasting}

\textbf{Activity-to-concept feedback.}
To enable interventions on the activity distribution, we learn direct activity-to-concept feedback.
TRACE represents this feedback with a sparse directed bipartite graph from $K$ activity classes to $C$ named concepts. 
Its rows correspond to activity classes, its columns to concepts, and $A_{kj}$ is the raw message strength from activity $k$ to concept $j$.
Its effective adjacency matrix is
\begin{equation}
    \mathbf{A}_{\mathrm{act}}
    =\mathbf{A}\odot\sigma(\boldsymbol{\lambda}_{A})
    \odot\mathbf{M}_{\mathrm{q_{\mathrm{fb}}}}
    \in\mathbb{R}^{K\times C},
    \label{eq:activity-feedback}
\end{equation}
where $\boldsymbol{\lambda}_{A}$ contains learned gate logits and
$\mathbf{M}_{\mathrm{q_{\mathrm{fb}}}}$ is a binary top-$\mathrm{q_{\mathrm{fb}}}$ mask that retains the strongest outgoing concept relations for each activity class. 
Given initial predicted activity probabilities $\hat{\mathbf y}_t$ (before feedback refinement), $\hat{\mathbf y}_t\mathbf{A}_{\mathrm{act}}$ is the activity-derived concept feedback. 
At window $t$, we aggregate this signal from the preceding $L_{\mathrm{hist}}$ windows:
\begin{equation}
\mathbf{d}^{\mathrm{hist}}_t
=
\sum_{\delta=1}^{L_{\mathrm{hist}}}
\sigma(\boldsymbol{\lambda}_{\delta})\,m_{t-\delta}\,
\hat{\mathbf y}_{t-\delta}\mathbf{A}_{\mathrm{act}}.
    \label{eq:historical-activity-feedback}
\end{equation}
where $\delta$ indexes the temporal lag, $\lambda_\delta$ is a learned lag gate, and $m_{t-\delta}\in\{0,1\}$ is a validity indicator that equals $1$ for an observed or already generated rollout state and $0$ for padding or
unavailable history.
This feedback refines the activity state and the forecast state:
\begin{equation}
    \widetilde{\mathbf{r}}^{\,\alpha}_t
    =
    \sigma\!\left(
        \phi(\mathbf{r}^{\alpha}_t)+\mathbf{d}^{\mathrm{hist}}_t
    \right),
    \qquad
    \alpha\in\{\mathrm{act},\mathrm{fut}\}.
    \label{eq:historical-feedback-update}
\end{equation}

For $t<T$, we set $\widetilde{\mathbf{r}}^{\,\alpha}_t=\mathbf{r}^{\alpha}_t$.
Thus, $\widetilde{\mathbf{r}}^{\,\alpha}_{t}$ denotes the refined concept state after activity feedback.

\textbf{Activity-conditioned autoregressive forecasting.}
The forecast module then recursively updates the concept state starting from $\widetilde{\mathbf{r}}^{\,\mathrm{fut}}_{1:T}$, with $\widetilde{\mathbf{r}}^{\,\mathrm{fut}}_{1:T+h}$ denoting the forecast at $T+h$:
\begin{equation}
    {\widetilde{\mathbf{r}}^{\,\mathrm{fut}}_{1:t+1}}
      =\sigma\!\left(
          \phi\!\left(\mathcal{G}_{\mathrm{fut}}\!\left(\widetilde{\mathbf{r}}^{\,\mathrm{fut}}_{1:t}\right)\right)
          + 
          \mathbf{d}^{\mathrm{hist}}_{t+1}
       \right).
    \label{eq:activity-conditioned-rollout}
\end{equation}

At each time step, feedback is applied conditioned on the history of past activity states $\mathbf{d}^{\mathrm{hist}}_t$.

\textbf{Shared activity head.}
Since we want to have a shared predictor that maps concept states to activity labels, both current activity recognition at time $T$ and future activity forecasting use the same linear predictor (Activity Head), parameterized by weights $\mathbf{W}\in\mathbb{R}^{K\times C}$ and bias $\mathbf{b}\in\mathbb{R}^{K}$:

\begin{equation}
    \boldsymbol\ell_t
    =\phi\!\left(\widetilde{\mathbf{r}}^{\,\alpha}_t\right)
       \mathbf{W}^\top+\mathbf{b},
    \qquad
    \hat{\mathbf{y}}_t=\operatorname{softmax}(\boldsymbol\ell_t),\quad m_t=1.
    \label{eq:current-prediction}
\end{equation}

Accordingly, the activity head predicts the activity at the future time $T+h$ from the forecasted concept state $\widetilde{\mathbf{r}}^{\,\mathrm{fut}}_{T+h}$.

\subsection{Training Objective}
\label{sec:optimization}

Future concept targets are obtained by applying the learned normalizer (Eq. \ref{eq:calibration}) to the CLIP-based concept scores of future states. 
The learned normalizer is excluded from gradient computation for these targets and, therefore, does not update its parameters from the target path and is instead trained through the observed-concept branch.
We train the model with a multi-task objective:
\begin{equation}
\mathcal{L}
=
\mathcal{L}_{\mathrm{act}}
+
\mathcal{L}_{\mathrm{forecast}}
+
\lambda_{\mathrm{c}}\mathcal{L}_{\mathrm{concept}}
+
\lambda_{\mathrm{cls}}\lVert \mathbf{W}\rVert_1 .
\label{eq:training-objective}
\end{equation}
Here, $\mathcal{L}_{\mathrm{act}}$ supervises the current activity prediction, $\mathcal{L}_{\mathrm{forecast}}$ supervises future activities across prediction horizons, and $\mathcal{L}_{\mathrm{concept}}$ aligns predicted concept states with the stop-gradient future concept targets. 
The $\ell_1$ penalty $\lambda_{\mathrm{cls}}\lVert \mathbf{W}\rVert_1$ promotes sparsity in the activity-classification weights.
The activity and forecasting terms use background-silence (SIL)-aware supervision. 
Forecasting additionally employs transition-tolerant targets for near-boundary predictions. 
Full loss definitions, horizon weights, and tolerance sets appear in Appendix~\ref{app:training-objective}.

\subsection{Interventions}
\label{sec:intervention_section}

TRACE’s interpretable design enables targeted investigations and interventions (edits) without changing its model parameters.
We support three types of interventions:
(i) \emph{What if a concept were different?}: replace or clamp a named concept activation in $[0,1]$.
(ii) \emph{How does changing an activity belief affect future concept states?}: modify a source activity distribution by setting one class to a chosen probability and redistributing the remaining mass.
(iii) \emph{Which relational dependencies does the TRACE use?}: modify a selected effective same-time, self-temporal, or cross-temporal relation while leaving the residual path unchanged.
Since all states remain in the same named concept space, predictions remain interpretable through the shared decoder, while the graphs expose the learned relations through which an intervention acts.
An edit at $T$ changes the initial concept state, activity belief, or relation for $h=1$, thereby affecting the prediction at $T+1$ (see the node intervention in Fig.~\ref{fig:motivation}).
The resulting state and activity distribution then feed into the $h=2$ transition, recursively propagating the intervention through the same autoregressive rollout up to $h=H$.

\section{Evaluation}
\label{sec:evaluation}

In this section, we address the research questions on TRACE’s intervenability introduced in Section \ref{sec:intervention_section}. 
We first describe the experimental setup, then present quantitative results in Section \ref{sec:quanti} and evaluate TRACE’s intervenability in Section \ref{sec:intervention-evaluation}. 
Additional experiments and ablations are provided in the Appendix.

\textbf{Datasets.}
We evaluate TRACE on three video activity datasets: Breakfast \citep{breakfast}, MPII Cooking~2 \citep{rohrbach15ijcv}, and BARISTA \citep{knab2026barista}. 
The main comparison uses BARISTA, averages the official Breakfast s1–s4 train-test splits, and uses MPII Cooking~2 (Attr). 
Ablations and intervention experiments use the same datasets and protocols, except that Breakfast is evaluated on the official s1 split only.
Further datasets are evaluated in Appendices \ref{app:gtea_epic} and \ref{app:synthetic-delayed}.
All methods use five observed 32-frame windows and forecast $H{=}3$ steps ahead.
We use a frozen PE-L14 \citep{bolya2025perception} backbone to obtain video and concept embeddings and action-label targets, with SIL assigned to otherwise unlabeled windows.
Appendix~\ref{app:dataset-baseline-details} gives the target definitions and complete protocol details, while Appendix~\ref{app:pe-g14-main-three} and Appendix~\ref{app:all-protocols} report the PE-G14 backbone.

\textbf{Concepts.}
For each dataset, we construct a $|C|{=}128$ textual concept vocabulary from training windows using a VLM-based pipeline to propose domain-specific concepts \citep{oikarinen2023labelfree,yang2023language}.
Qwen3.5-27B \citep{qwenteam2026qwen35omnitechnicalreport} receives four frames of a window, and its proposed concepts are retrieved, visually verified, deduplicated, and filtered for label overlap before training (see Appendix~\ref{app:vlm-concepts}).

\textbf{Baselines.}
We compare TRACE against a training-free zero-shot baseline (PE-L/14). 
For black-box models, we use SlowFast-style and feature-transformer architectures~\citep{feichtenhofer2019slowfast,bertasius2021space}. 
As concept-based approaches, we use a Sparse Linear baseline with matched concept-space sparsity and a decoder that replaces the graph component, and MoTIF~\citep{knab2026concepts} for activity recognition only.
The zero-shot baseline predicts the activity with cosine similarity between its text prompt ``a video of \textit{label}’’ and the frozen video-window feature \citep{bolya2025perception}.
The black-box models operate on the same frozen video features without concept activations or graph structure, Appendix~\ref{app:implementation} provides the implementation and complete configurations.

\subsection{Quantitative evaluation}
\label{sec:quanti}

\textbf{Concept-space forecasting remains competitive with black-box baselines.}
\label{eval:main_comparison}
Table~\ref{tab:blackbox-concept-baselines} shows that TRACE achieves the strongest activity accuracy among concept-based methods across all three datasets and substantially narrows the gap to black-box feature baselines, while additionally providing interpretability and intervention capabilities (Section~\ref{sec:intervention-evaluation}).
Among concept-based methods, our model improves activity top-1 accuracy over MoTIF on all three datasets and remains the overall best on H3 forecasting, while Sparse Linear performs comparably by capturing direct concept-to-concept dependencies through its sparse coefficient matrix but lacks structured concept/activity intervention mechanisms.
The substantially weaker zero-shot baseline further highlights the importance of learning over concept representations.
Thus, TRACE remains competitive with black-box models while exposing internal semantic states that can be investigated and intervened on.
On H3, TRACE is at parity with Sparse Linear, and black-box models remain slightly stronger on windows where the activity changes (Appendix~\ref{app:change-forecast}).

\begin{table}[t]
\centering
\caption{\textbf{Predictive performance across model families.} Mean $\pm$ sample standard deviation over five seeds for activity classification and H3 forecasting on BARISTA, Breakfast s1--s4 (averaged within seed), and MPII Attr. MoTIF reports classification only; black-box baselines use the same window features without concept activations.}
\label{tab:blackbox-concept-baselines}
\begingroup
\scriptsize
\setlength{\tabcolsep}{2pt}
\resizebox{\textwidth}{!}{%
\begin{tabular}{llcccccccccccc}
\toprule
& & \multicolumn{4}{c}{BARISTA} &
\multicolumn{4}{c}{Breakfast (s1--s4)} &
\multicolumn{4}{c}{MPII Cooking~2 (Attr)} \\
\cmidrule(lr){3-6} \cmidrule(lr){7-10} \cmidrule(lr){11-14}
& & \multicolumn{2}{c}{Activity} & \multicolumn{2}{c}{H3}
& \multicolumn{2}{c}{Activity} & \multicolumn{2}{c}{H3}
& \multicolumn{2}{c}{Activity} & \multicolumn{2}{c}{H3} \\
\cmidrule(lr){3-4} \cmidrule(lr){5-6}
\cmidrule(lr){7-8} \cmidrule(lr){9-10}
\cmidrule(lr){11-12} \cmidrule(lr){13-14}
Family & Method & Top-1 & Top-3 & Top-1 & Top-3
& Top-1 & Top-3 & Top-1 & Top-3
& Top-1 & Top-3 & Top-1 & Top-3 \\
\midrule

Zero-shot
& Cos. sim.
& 2.2 & 5.4 & n/a & n/a
& $\mval{18.2}{2.8}$ & $\mval{37.3}{4.7}$ & n/a & n/a
& 6.8 & 15.9 & n/a & n/a \\

\midrule

\multirow{2}{*}{\shortstack[l]{Black-box}}
& Feat. Transf.
& $\mval{79.2}{0.3}$ & $\mval{89.3}{0.5}$
& $\mval{69.6}{0.3}$ & $\mval{82.9}{0.3}$
& $\mval{47.3}{0.4}$ & $\mval{69.6}{0.6}$
& $\mval{39.7}{0.6}$ & $\mval{63.7}{0.2}$
& $\mval{50.1}{1.2}$ & $\mval{70.3}{1.4}$
& $\mval{40.7}{0.7}$ & $\mval{61.7}{0.8}$ \\

& SlowFast
& $\mval{79.6}{0.5}$ & $\mval{90.3}{0.9}$
& $\mval{67.6}{1.0}$ & $\mval{82.5}{0.8}$
& $\mval{46.8}{0.5}$ & $\mval{69.5}{0.6}$
& $\mval{39.7}{0.6}$ & $\mval{64.1}{0.4}$
& $\mval{49.8}{0.6}$ & $\mval{70.4}{0.5}$
& $\mval{40.4}{0.4}$ & $\mval{62.3}{0.5}$ \\

\midrule

\multirow{3}{*}{\shortstack[l]{Concept}}
& Sparse Lin.
& $\mval{67.5}{1.4}$ & $\mval{79.3}{1.5}$
& $\bmval{68.5}{0.4}$
& $\bmval{82.5}{0.3}$
& $\mval{43.4}{0.2}$ & $\mval{67.9}{0.1}$
& \underline{$\mval{36.8}{0.2}$}
& \underline{$\mval{61.5}{0.2}$}
& $\mval{43.3}{0.3}$ & $\mval{61.5}{0.5}$
& \underline{$\mval{37.7}{0.2}$}
& \underline{$\mval{58.3}{0.3}$} \\

& MoTIF
& \underline{$\mval{72.9}{1.9}$}
& \underline{$\mval{86.2}{0.8}$}
& n/a & n/a
& \underline{$\mval{44.3}{0.3}$}
& \underline{$\mval{69.9}{0.3}$}
& n/a & n/a
& \underline{$\mval{44.4}{0.9}$}
& \underline{$\mval{65.6}{0.7}$}
& n/a & n/a \\

& TRACE (ours)
& $\bmval{76.0}{2.2}$
& $\bmval{88.9}{0.8}$
& \underline{$\mval{67.6}{0.3}$}
& \underline{$\mval{81.5}{1.1}$}
& $\bmval{45.5}{0.2}$
& $\bmval{70.6}{0.2}$
& $\bmval{37.5}{0.1}$
& $\bmval{62.7}{0.4}$
& $\bmval{47.0}{0.4}$
& $\bmval{67.4}{0.2}$
& $\bmval{38.8}{0.5}$
& $\bmval{58.6}{0.3}$ \\

\bottomrule
\end{tabular}%
}
\endgroup
\end{table}

\begin{wraptable}[11]{r}{0.56\textwidth}
    \vspace{-0.45cm}
    \centering
    \caption{\textbf{Component ablation.} H3 top-1 accuracy, reported as mean $\pm$ sample standard deviation over three seeds. Activity and H3 top-3 results are in Appendix~\ref{app:transition-factorial}.}
    \label{tab:component-ablation}
    \begingroup
    \scriptsize
    \setlength{\tabcolsep}{2pt}
    \resizebox{\linewidth}{!}{%
    \begin{tabular}{lccc}
        \toprule
        Variant & BARISTA & Breakfast & MPII Attr \\
        \midrule
        \textbf{Full model} & \underline{$\mval{67.4}{0.3}$} & $\bmval{42.9}{0.2}$ & $\bmval{38.7}{0.6}$ \\
        Persistence & $\mval{66.0}{0.2}$ & $\bmval{42.9}{0.1}$ & \underline{$\mval{38.2}{0.3}$} \\
        Dense rollout, no graph & $\bmval{70.2}{0.3}$ & $\mval{40.2}{0.3}$ & $\mval{35.9}{0.3}$ \\
        Graph, no activity feedback & $\mval{67.0}{0.1}$ & \underline{$\mval{42.5}{0.6}$} & $\bmval{38.7}{0.4}$ \\
        No score normalization & $\mval{64.3}{0.5}$ & $\mval{39.6}{0.7}$ & $\mval{35.0}{0.3}$ \\
        \bottomrule
    \end{tabular}%
    }
    \endgroup
\end{wraptable}

\textbf{Component ablation.}
\label{sec:component-ablation}
Here, we ablate whether the full model benefits from all of its components.
Table~\ref{tab:component-ablation} separately removes recursive transitions, graph structure, activity feedback, and score normalization on BARISTA, Breakfast s1, and MPII Cooking 2.
The complete model achieves the highest mean H3 top-1 accuracy across the three datasets ($49.6\%$), while score normalization consistently helps to improve forecasting performance.
Dense instead of graph rollout improves BARISTA but reduces Breakfast and MPII performance by 2.7 and 2.8 points, respectively.
However, as shown by the dependency-concentration metric in Appendix~\ref{app:jacobian-dependency}, the dense rollout is substantially less sparse (~87 effective H3 parents vs. ~5 for graph rollout), and the weight of concept relations is less concentrated (~15\% influence of the top 5 concepts compared to 75\% for graph rollout).
While the sparsity of the graph rollout is generally desirable, TRACE can also operate with the dense rollout when desired.
Further ablations in Appendix~\ref{app:additional-quantitative}.

\subsection{Intervention evaluation}
\label{sec:intervention-evaluation}

\begin{figure*}[t]
    \centering
    \includegraphics[width=\textwidth]{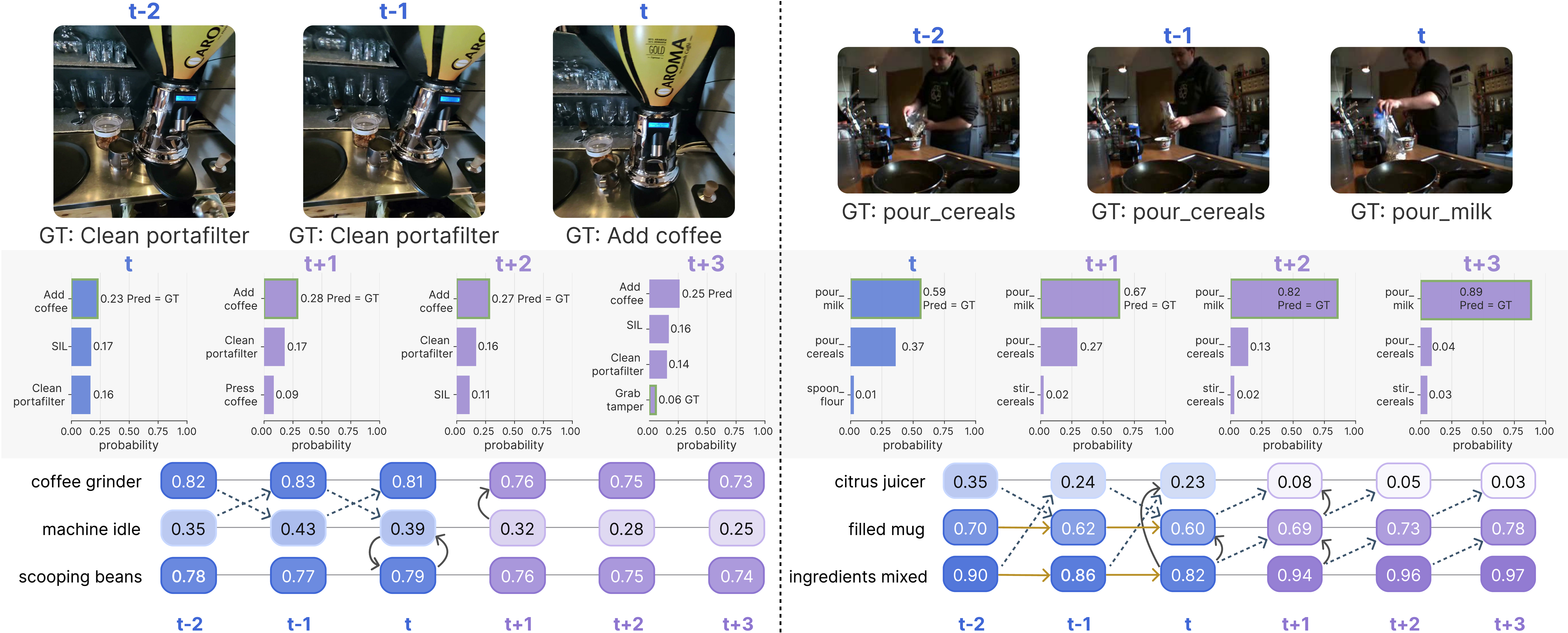}
    \caption{\textbf{Exemplary forecasts.} Representative BARISTA and Breakfast sequences show observed windows, activity distributions decoded from refined and forecasted concept states, and selected learned relations used by the model. }
    \label{fig:qualitative-forecasting}
\end{figure*}

\textbf{What are the predictive effects of interventions?}
In Table~\ref{tab:intervention-interfaces} we evaluate the intervenability with up to a budget of $b = 5$ cumulative edits on 30 initially wrong (W) and correct (C) cases per dataset, recomputing the full rollout after each edit, answering questions (i) and (ii) in Section \ref{sec:intervention_section}.
Activity edits use ground-truth activity with $b = 3$, node edits use future next-window (H1) concept targets and the true H3 label for ranking, edge edits flip the strongest learned relations without future labels.
This extends ground-truth (\textit{oracle}) test-time concept interventions \citep{koh2020concept,shin2023intervention, zarlenga2023intcem} to forecasting. 
Activity-belief and concept-node edits both improve H3 forecasts. However, while activity-belief edits correct more cases on Breakfast and MPII node edits consistently increase true-class probability without inducing misclassifications. 
The negligible effect of single-edge flips indicates robustness to individual relational perturbations, meaning the edges serve more for inspection (see Figure \ref{fig:qualitative-forecasting}).
Together, these results demonstrate that TRACE can be corrected through interventions on activity-conditioned transitions and named concept states.
Final rates, the information available to each policy, and matched editable baselines are in Appendix~\ref{app:final-rates}.
\begin{table*}[t]
    \centering
    \caption{\textbf{Model edits for level H3.} Interventions on 30 wrong and correct cases per seed, with activity-belief ($b{=}3$), concept-node ($b{=}5$), and strongest edge sign edits ($b{=}5$). 
    $\Delta p(y)$ is the true-class probability change (\%); W$\rightarrow$C and C$\rightarrow$W are cumulative correction and harm rates.}
    \label{tab:intervention-interfaces}
    \begingroup
    \scriptsize
    \setlength{\tabcolsep}{2.5pt}
    \resizebox{\textwidth}{!}{%
    \begin{tabular}{lrrrrrrrrr}
        \toprule
        & \multicolumn{3}{c}{Activity belief (A, $b{=}3$)} & \multicolumn{3}{c}{Concept node (N, $b{=}5$)} & \multicolumn{3}{c}{Edge sign flip (E, $b{=}5$)} \\
        \cmidrule(lr){2-4} \cmidrule(lr){5-7} \cmidrule(lr){8-10}
        Dataset & $\Delta p(y)$ & W$\rightarrow$C & C$\rightarrow$W & $\Delta p(y)$ & W$\rightarrow$C & C$\rightarrow$W & $\Delta p(y)$ & W$\rightarrow$C & C$\rightarrow$W \\
        \midrule
        BARISTA & $\mval{+3.0}{0.5}$ & $\mval{12.2}{8.4}$ & $\mval{0.0}{0.0}$ & $\mval{+7.2}{5.3}$ & $\mval{25.6}{18.4}$ & $\mval{0.0}{0.0}$ & $\mval{-0.02}{0.02}$ & $\mval{0.0}{0.0}$ & $\mval{1.1}{1.9}$ \\
        Breakfast s1 & $\mval{+28.3}{4.2}$ & $\mval{61.1}{5.1}$ & $\mval{4.4}{5.1}$ & $\mval{+11.3}{1.2}$ & $\mval{25.6}{3.8}$ & $\mval{0.0}{0.0}$ & $\mval{-0.00}{0.11}$ & $\mval{0.0}{0.0}$ & $\mval{0.0}{0.0}$ \\
        MPII Attr & $\mval{+21.1}{1.7}$ & $\mval{46.7}{8.8}$ & $\mval{2.2}{3.8}$ & $\mval{+9.8}{1.5}$ & $\mval{34.4}{6.9}$ & $\mval{0.0}{0.0}$ & $\mval{+0.04}{0.05}$ & $\mval{2.2}{1.9}$ & $\mval{1.1}{1.9}$ \\
        \bottomrule
    \end{tabular}%
    }
    \endgroup
\end{table*}

\textbf{Can a VLM plan concept edits?}
TRACE functionality allows to test whether a VLM, Qwen3.6-35B-A3B \citep{qwenteam2026qwen35omnitechnicalreport}, can act as an external intervention agent for the observed window $T$ and forecast $T{+}1$ and $T{+}2$. 
For this, we construct four different test scenarios: blind Qwen receives the model state and candidate edits, while oracle-assisted variants additionally receive the true H3 label and counterfactual edit effects. 
We compare one-shot intervention, sequential intervention with updated effects ($b = 5$), and a greedy oracle that reruns each next edit and selects the one maximizing held-out true H3 probability in Table~\ref{tab:qwen-intervention-planning} (Appendix~\ref{app:qwen-prompt} provides the full setup and prompt).

\begin{table}[h]
    \centering
    \caption{\textbf{Qwen planning for H3 concept-node interventions.} Mean $\pm$ standard deviation after five cumulative concept-node edits over three seeds, with 30 initially wrong and 30 correct cases per dataset and seed. $\Delta p(y)$ is in percentage points; W$\rightarrow$C is the cumulative correction rate. Blind Qwen receives neither the future label nor counterfactual effects; one-shot and sequential are oracle-assisted.}
    \label{tab:qwen-intervention-planning}
    \begingroup
    \scriptsize
    \setlength{\tabcolsep}{2.5pt}
    \resizebox{\textwidth}{!}{%
    \begin{tabular}{lrrrrrrrr}
        \toprule
        & \multicolumn{2}{c}{Blind Qwen} & \multicolumn{2}{c}{One-shot Qwen} & \multicolumn{2}{c}{Sequential Qwen} & \multicolumn{2}{c}{Greedy oracle} \\
        \cmidrule(lr){2-3} \cmidrule(lr){4-5} \cmidrule(lr){6-7} \cmidrule(lr){8-9}
        Dataset & $\Delta p(y)$ & W$\rightarrow$C & $\Delta p(y)$ & W$\rightarrow$C & $\Delta p(y)$ & W$\rightarrow$C & $\Delta p(y)$ & W$\rightarrow$C \\
        \midrule
        BARISTA & $\mval{-2.9}{1.7}$ & $\mval{10.0}{8.8}$ & $\mval{+12.9}{4.7}$ & {$\mval{47.8}{13.9}$} & \underline{$\mval{+16.6}{6.1}$} & $\bmval{57.8}{10.7}$ & $\bmval{+17.8}{6.8}$ & $\bmval{57.8}{16.4}$ \\
        Breakfast s1 & $\mval{-13.0}{5.7}$ & $\mval{13.3}{6.7}$ & $\mval{+23.1}{3.2}$ & {$\mval{56.7}{12.0}$} & \underline{$\mval{+32.4}{4.7}$} & $\bmval{75.6}{3.8}$ & $\bmval{+34.4}{5.2}$ & \underline{$\mval{72.2}{5.1}$} \\
        MPII Attr & $\mval{-9.0}{1.6}$ & $\mval{12.2}{3.8}$ & $\mval{+20.6}{2.4}$ & {$\mval{52.2}{10.7}$} & \underline{$\mval{+25.7}{3.3}$} & \underline{$\mval{58.9}{7.7}$} & $\bmval{+28.5}{4.0}$ & $\bmval{61.1}{9.6}$ \\
        \bottomrule
    \end{tabular}%
    }
    \endgroup
\end{table}
\begin{wraptable}[14]{r}{0.5\textwidth}
    \vspace{-0.4cm}
    \centering
    \caption{\textbf{H3 steering}. Mean $\pm$ sample standard deviation over three seeds with 60 test windows per dataset and seed. $\Delta p(a^\star)$ is in percentage points; $p\uparrow$ is the target-increase rate.}
    \label{tab:non-oracle-steering}
    \begingroup
    \setlength{\tabcolsep}{2pt}
    \resizebox{\linewidth}{!}{%
    \begin{tabular}{llrrr}
        \toprule
        Dataset & Policy & $\Delta p(a^\star)$ & $p\uparrow$ & Target top-1 \\
        \midrule
        BARISTA
        & Random & $\mval{-0.4}{0.7}$ & $\mval{41.7}{7.3}$ & $\mval{1.7}{1.7}$ \\
        & Target heuristic & $\mval{+4.0}{1.5}$ & $\mval{80.0}{10.9}$ & $\mval{3.9}{4.2}$ \\
        & Qwen steering & $\mval{+5.8}{3.3}$ & $\mval{87.2}{14.9}$ & $\mval{10.6}{8.6}$ \\
        \addlinespace
        Breakfast s1
        & Random & $\mval{+3.7}{2.9}$ & $\mval{54.4}{7.5}$ & $\mval{11.7}{1.7}$ \\
        & Target heuristic & $\mval{+68.9}{1.6}$ & $\mval{99.4}{1.0}$ & $\mval{90.0}{3.3}$ \\
        & Qwen steering & $\mval{+64.0}{3.3}$ & $\mval{98.9}{1.0}$ & $\mval{88.9}{2.5}$ \\
        \addlinespace
        MPII Attr
        & Random & $\mval{+2.0}{0.7}$ & $\mval{43.3}{3.3}$ & $\mval{8.3}{0.0}$ \\
        & Target heuristic & $\mval{+51.9}{6.5}$ & $\mval{100.0}{0.0}$ & $\mval{84.4}{8.6}$ \\
        & Qwen steering & $\mval{+47.7}{4.4}$ & $\mval{98.3}{1.7}$ & $\mval{78.3}{1.7}$ \\
        \bottomrule
    \end{tabular}%
    }
    \endgroup
\end{wraptable}
Blind Qwen corrects some cases, but overall lowers $p(y)$.
Oracle information makes one-shot and sequential planning effective, with sequential being close to greedy. 
Together, these results show that Qwen can plan corrective concept edits when provided with information about the future label and counterfactual edit effects. 
The next experiment tests a related non-oracle setting by asking Qwen to steer the model toward a specified target activity.

\textbf{Can a VLM steer a requested H3 target?}
We select a requested activity from the three most likely train-only H3 transitions under the model's current activity belief, excluding its default prediction. 
For each case, all policies receive on/off edits for 10 concepts selected by their contribution to the requested activity and activity-belief edits for three candidates, including the target. 
Random shuffles these edits, the target heuristic prioritizes target-activity and then concept-on edits, and Qwen selects five nonconflicting cumulative edits from the observed storyboard and model state. 
Future labels, and intervention effects remain hidden (Appendix~\ref{app:qwen-prompt}).
Both steering strategies outperform the random baseline (Table~\ref{tab:non-oracle-steering}). 
Although Qwen consistently increases the requested activity’s probability, the target heuristic is stronger on Breakfast and MPII. 
BARISTA also exhibits positive target gains, but its activity and concept interventions have more distributed effects.
The strongest influence comes from activity-belief edits (Appendix~\ref{app:steering-split}).


\textbf{Which relational dependencies does TRACE use?}
Figure~\ref{fig:qualitative-forecasting} visualizes representative BARISTA and Breakfast sequences using the same backbone as the quantitative evaluation.
The figure shows the information flow of TRACE: observed windows, activity distributions decoded from refined and forecasted concept states, and learned graph relations used for state refinement.
By inspecting these relations and concept scores, we can answer: \emph{Which relational dependencies does the forecaster use?}
Unlike an opaque latent trajectory, future predictions remain readable through named concept states, activity beliefs, and graph dependencies.
Each example displays a subset of the most relevant concept nodes and their learned relations (we omit edge strengths for visual clarity).
As motivated in Section~\ref{sec:method}, the sparse graphs connect only selected concept pairs rather than imposing dependencies between all concepts.
This sparsity still lets us \textit{trace} how interventions propagate through learned relations and, for BARISTA, why the model falsely predicts \textit{add coffee} at $t{+}3$.

\textbf{What drives TRACE's predictions?}
Complementing the previous intervention studies and the examples shown in Figure \ref{fig:qualitative-forecasting}, we now assess the interpretability of TRACE’s concept-based predictions through targeted manual interventions on influential concepts and activities.
For BARISTA, suppressing \textit{coffee grinder} to $0.1$ changes the preferred activity from \textit{add coffee} to \textit{SIL} for H2 and H3. 
For Breakfast, suppressing \textit{filled mug} lowers $p$ for \textit{pour\_milk} and increases \textit{pour\_cereals}. 
After a second intervention by suppressing \textit{ingredients mixed}, TRACE predicts the activity \textit{pour\_cereals} for H2 and H3. 
These interventions show that we humans can steer its predictions through the concept space and why TRACE prefers its original prediction over a plausible alternative.

\section{Conclusion and future work}
\label{sec:conclusion}

\textbf{Conclusion.}
We introduced TRACE, a concept-rollout model for intervenable video activity forecasting.
The model refines observed concept evidence using sparse spatio-temporal graph modules and recursively forecasts named concept states with activity-conditioned feedback.
This shared semantic state supports retraining-free edits to concepts, activity beliefs, and learned edges, while sparse graph dynamics preserve localized, inspectable concept contributions throughout forecasting.
TRACE therefore provides an editable interface beyond the observed input and across the forecast horizon.

\textbf{Limitations.}
Although this interface makes forecasts easier to inspect and modify, several limitations remain.
The frozen video encoder and fixed 32-frame windows may miss the task-specific appearance or fine temporal changes, adapted representations and variable temporal context could improve both.
Text--video-aligned concepts can introduce label noise, and direct concept supervision would enable a stronger evaluation of grounding and leakage \citep{zarlenga2022concept} (a VLM and human audit of the predicted states is in Appendix~\ref{app:concept-grounding}).
The exact learned edges between concepts vary across seeds, although shared edges usually retain their direction. 
The graphs should therefore be read as predictive dependencies rather than verified causal structure, and selecting concept edits without future information remains open.

\textbf{Future work.} 
Promising directions include adaptive concept structures, longer-horizon and uncertainty-aware forecasting, and transfer to multivariate time series.
Our interventions expose mechanisms for modifying forecasts, but a deployable policy to detect unreliable concepts and select corrections without future labels remains open \citep{steinmann2024intervene,zarlenga2023intcem}.
Improving forecasting accuracy while retaining interpretability and intervenability is another important direction.
Finally, extending persistent concept interventions from image generation \citep{ismail2024conceptbottleneckgenerative,martinezgarcia2026probabilistichardconcept} to generative video models could enable semantic edits that remain consistent across generated frames \citep{labs2025flux1kontext}.




\subsubsection*{Acknowledgments}
This research was supported in part by the German Federal Ministry for Economic Affairs and Climate Action of Germany (BMWK) and by the German Federal Ministry for Research, Technology, and Space (BMFTR).
GDF acknowledges support from the Swiss National Science Foundation (SNSF) through the grant 205121\_197242 for the project “PROSELF: Semi-automated Self-Tracking Systems to Improve Personal Productivity”, and by the Hasler Foundation through the project “Towards Scalable Multimodal Causal Deep Learning” (grant ID 2024-05-15-70).
The project was further supported by the ”ML2MT” project from the Volkswagen Stiftung, by the German Research Foundation (DFG) under Germany’s Excellence Strategy (EXC 3066/1 “The Adaptive Mind”, Project No. 533717223 and by the EU-funded “TANGO” project (EU Horizon 2023, GA No 57100431). 
It has benefited from the HMWK project Hessian.AI, the Cluster of Excellence "Reasonable AI" funded by the DFG under Germany’s Excellence Strategy EXC-3057 and the GRK 2853 "Neuroexplicit Models of Language, Vision, and Action" (Project No. 471607914).

\bibliography{references}

@InProceedings{koh2020concept,
  title = 	 {Concept Bottleneck Models},
  author =       {Koh, Pang Wei and Nguyen, Thao and Tang, Yew Siang and Mussmann, Stephen and Pierson, Emma and Kim, Been and Liang, Percy},
  booktitle = 	 {Proceedings of the 37th International Conference on Machine Learning},
  pages = 	 {5338--5348},
  year = 	 {2020},
  editor = 	 {III, Hal Daumé and Singh, Aarti},
  volume = 	 {119},
  series = 	 {Proceedings of Machine Learning Research},
  month = 	 {13--18 Jul},
  publisher =    {PMLR}
}

@article{zarlenga2022concept,
  title={Concept embedding models: Beyond the accuracy-explainability trade-off},
  author={Espinosa Zarlenga, Mateo and Barbiero, Pietro and Ciravegna, Gabriele and Marra, Giuseppe and Giannini, Francesco and Diligenti, Michelangelo and Shams, Zohreh and Precioso, Frederic and Melacci, Stefano and Weller, Adrian and others},
  journal={Advances in neural information processing systems},
  volume={35},
  pages={21400--21413},
  year={2022}
}

@inproceedings{shin2023intervention,
  title={A closer look at the intervention procedure of concept bottleneck models},
  author={Shin, Sungbin and Jo, Yohan and Ahn, Sungsoo and Lee, Namhoon},
  booktitle={International conference on machine learning},
  pages={31504--31520},
  year={2023},
  organization={PMLR}
}

@inproceedings{
steinmann2024intervene,
title={Learning to Intervene on Concept Bottlenecks},
author={David Steinmann and Wolfgang Stammer and Felix Friedrich and Kristian Kersting},
booktitle={Forty-first International Conference on Machine Learning},
year={2024},
}

@inproceedings{zarlenga2023intcem,
  title={Learning to Receive Help: Intervention-Aware Concept Embedding Models},
  author={Espinosa Zarlenga, Mateo and Collins, Katherine M. and Dvijotham, Krishnamurthy and Weller, Adrian and Shams, Zohreh and Jamnik, Mateja},
  booktitle={Advances in Neural Information Processing Systems},
  volume={36},
  year={2023},
}

@inproceedings{
felice2026position,
title={Position: Interpretability in Deep Time Series Models Demands Semantic Alignment},
author={De Felice, Giovanni and D'Elia, Riccardo Giuseppe and Termine, Alberto and Barbiero, Pietro and Marra, Giuseppe and Santini, Silvia},
booktitle={Forty-third International Conference on Machine Learning Position Paper Track},
year={2026},
}

@inproceedings{
knab2026concepts,
title={Concepts in Motion: Temporal Concept Bottleneck Model for Interpretable Video Classification},
author={Patrick Knab and Sascha Marton and Philipp Johannes Schubert and Drago Andres Guggiana Nilo and Christian Bartelt},
booktitle={2nd Workshop on Compositional Learning: Safety, Interpretability, and Agents},
year={2026},
}

@inproceedings{
knab2026barista,
title={{BARISTA}: A Multi-Task Egocentric Benchmark for Compositional Visual Understanding},
author={Patrick Knab and Orgest Xhelili and Inis Buzi and Drago Andres Guggiana Nilo and Mohd Saquib Khan and Lorenz Kolb and Manuel Scherzer and Kerem Yildirir and Christian Bartelt and Philipp Johannes Schubert},
booktitle={ICML 2026 Workshop on Combining Theory and Benchmarks: Towards A Virtuous Cycle to Understand and Guarantee Foundation Model Performance},
year={2026},
}

@article{
knab2026whats,
title={What{\textquoteright}s in the Bottle? A Survey and Roadmap of Concept Bottleneck Models},
author={Patrick Knab and David Steinmann and Christian Bartelt and Kristian Kersting and Bernt Schiele and Thomas Seidl and Udo Schlegel and Wolfgang Stammer},
journal={Transactions on Machine Learning Research},
issn={2835-8856},
year={2026},
note={}
}

@inproceedings{assran2023self,
  title={Self-supervised learning from images with a joint-embedding predictive architecture},
  author={Assran, Mahmoud and Duval, Quentin and Misra, Ishan and Bojanowski, Piotr and Vincent, Pascal and Rabbat, Michael and LeCun, Yann and Ballas, Nicolas},
  booktitle={2023 IEEE/CVF Conference on Computer Vision and Pattern Recognition (CVPR)},
  pages={15619--15629},
  year={2023},
  organization={IEEE}
}

@inproceedings{
steinmann2026objectcentric,
title={Object-Centric Concept-Bottlenecks},
author={David Steinmann and Wolfgang Stammer and Antonia W{\"u}st and Kristian Kersting},
booktitle={The Thirty-ninth Annual Conference on Neural Information Processing Systems},
year={2025},
}

@inproceedings{
prasse2025dcbm,
title={{DCBM}: Data-Efficient Visual Concept Bottleneck Models},
author={Katharina Prasse and Patrick Knab and Sascha Marton and Christian Bartelt and Margret Keuper},
booktitle={Forty-second International Conference on Machine Learning},
year={2025},
}

@article{de2026causally,
  title={Causally reliable concept bottleneck models},
  author={De Felice, Giovanni and Casanova Flores, Arianna and De Santis, Francesco and Santini, Silvia and Schneider, Johannes and Barbiero, Pietro and Termine, Alberto},
  journal={Advances in Neural Information Processing Systems},
  volume={38},
  pages={149099--149139},
  year={2025}
}

@InProceedings{breakfast,
   author= "Kuehne, H. and Arslan, A. B. and Serre, T.",
   title = "The Language of Actions: Recovering the Syntax and Semantics of Goal-Directed Human Activities",
   booktitle = "Proceedings of Computer Vision and Pattern Recognition Conference (CVPR)",
   year = "2014",
}

@inproceedings{girdhar2021anticipative,
  title={Anticipative video transformer},
  author={Girdhar, Rohit and Grauman, Kristen},
  booktitle={Proceedings of the IEEE/CVF international conference on computer vision},
  pages={13505--13515},
  year={2021}
}

@inproceedings{gong2022future,
  title={Future transformer for long-term action anticipation},
  author={Gong, Dayoung and Lee, Joonseok and Kim, Manjin and Ha, Seong Jong and Cho, Minsu},
  booktitle={Proceedings of the IEEE/CVF Conference on Computer Vision and Pattern Recognition},
  pages={3052--3061},
  year={2022}
}

@inproceedings{liu2022video,
  title={Video swin transformer},
  author={Liu, Ze and Ning, Jia and Cao, Yue and Wei, Yixuan and Zhang, Zheng and Lin, Stephen and Hu, Han},
  booktitle={Proceedings of the IEEE/CVF conference on computer vision and pattern recognition},
  pages={3202--3211},
  year={2022}
}

@InProceedings{bertasius2021space,
  title = 	 {Is Space-Time Attention All You Need for Video Understanding?},
  author =       {Bertasius, Gedas and Wang, Heng and Torresani, Lorenzo},
  booktitle = 	 {Proceedings of the 38th International Conference on Machine Learning},
  pages = 	 {813--824},
  year = 	 {2021},
  editor = 	 {Meila, Marina and Zhang, Tong},
  volume = 	 {139},
  series = 	 {Proceedings of Machine Learning Research},
  month = 	 {18--24 Jul},
  publisher =    {PMLR}
}

@inproceedings{
dominici2025causal,
title={Causal Concept Graph Models: Beyond Causal Opacity in Deep Learning},
author={Gabriele Dominici and Pietro Barbiero and Mateo Espinosa Zarlenga and Alberto Termine and Martin Gjoreski and Giuseppe Marra and Marc Langheinrich},
booktitle={The Thirteenth International Conference on Learning Representations},
year={2025},
}

@inproceedings{feichtenhofer2019slowfast,
  title={Slowfast networks for video recognition},
  author={Feichtenhofer, Christoph and Fan, Haoqi and Malik, Jitendra and He, Kaiming},
  booktitle={Proceedings of the IEEE/CVF international conference on computer vision},
  pages={6202--6211},
  year={2019}
}

@inproceedings{yang2023language,
  title={Language in a bottle: Language model guided concept bottlenecks for interpretable image classification},
  author={Yang, Yue and Panagopoulou, Artemis and Zhou, Shenghao and Jin, Daniel and Callison-Burch, Chris and Yatskar, Mark},
  booktitle={Proceedings of the IEEE/CVF conference on computer vision and pattern recognition},
  pages={19187--19197},
  year={2023}
}

@inproceedings{
yuksekgonul2022posthoc,
title={Post-hoc Concept Bottleneck Models},
author={Mert Yuksekgonul and Maggie Wang and James Zou},
booktitle={ICLR 2022 Workshop on PAIR{\textasciicircum}2Struct: Privacy, Accountability, Interpretability, Robustness, Reasoning on Structured Data},
year={2022},
}

@inproceedings{
oikarinen2023labelfree,
title={Label-free Concept Bottleneck Models},
author={Tuomas Oikarinen and Subhro Das and Lam M. Nguyen and Tsui-Wei Weng},
booktitle={The Eleventh International Conference on Learning Representations },
year={2023},
}

@article{li2021eye,
  title={In the eye of the beholder: Gaze and actions in first person video},
  author={Li, Yin and Liu, Miao and Rehg, James M},
  journal={IEEE transactions on pattern analysis and machine intelligence},
  volume={45},
  number={6},
  pages={6731--6747},
  year={2021},
  publisher={IEEE}
}

@article{rohrbach15ijcv, 
year={2015}, 
journal={International Journal of Computer Vision}, 
title={Recognizing Fine-Grained and Composite Activities Using Hand-Centric Features and Script Data}, 
author={Rohrbach, Marcus and Rohrbach, Anna and Regneri, Michaela and Amin, Sikandar and Andriluka, Mykhaylo and Pinkal, Manfred and Schiele, Bernt}, 
issn={0920-5691}, doi={10.1007/s11263-015-0851-8}, publisher={Springer US}, pages={1-28} }

@article{damen2020epic,
  title={The epic-kitchens dataset: Collection, challenges and baselines},
  author={Damen, Dima and Doughty, Hazel and Farinella, Giovanni Maria and Fidler, Sanja and Furnari, Antonino and Kazakos, Evangelos and Moltisanti, Davide and Munro, Jonathan and Perrett, Toby and Price, Will and others},
  journal={IEEE Transactions on Pattern Analysis and Machine Intelligence},
  volume={43},
  number={11},
  pages={4125--4141},
  year={2020},
  publisher={IEEE}
}

@article{
xu2026graph,
title={Graph Concept Bottleneck Models},
author={Haotian Xu and Tsui-Wei Weng and Lam M. Nguyen and Tengfei Ma},
journal={Transactions on Machine Learning Research},
issn={2835-8856},
year={2026},
note={}
}

@inproceedings{
lee2025disentangled,
title={Disentangled Concepts Speak Louder Than Words: Explainable Video Action Recognition},
author={Jongseo Lee and Wooil Lee and Gyeong-Moon Park and Seong Tae Kim and Jinwoo Choi},
booktitle={The Thirty-ninth Annual Conference on Neural Information Processing Systems},
year={2025},
}

@inproceedings{yan2018spatial,
  title={Spatial temporal graph convolutional networks for skeleton-based action recognition},
  author={Yan, Sijie and Xiong, Yuanjun and Lin, Dahua},
  booktitle={Proceedings of the AAAI conference on artificial intelligence},
  volume={32},
  number={1},
  year={2018}
}

@misc{qwenteam2026qwen35omnitechnicalreport,
      title={Qwen3.5-Omni Technical Report}, 
      author={Qwen Team},
      year={2026},
      eprint={2604.15804},
      archivePrefix={arXiv},
      primaryClass={cs.CL},
}

@misc{sui2026recbmuncertaintygatedrelationalreasoning,
  title={ReCBM: Uncertainty-Gated Relational Reasoning for Concept Bottleneck Models},
  author={Sui, An and Li, Yuzhu and Wu, Fuping and Zhuang, Xiahai},
  journal={arXiv preprint arXiv:2608.10004},
  year={2026}
}

@InProceedings{clip,
  title = 	 {Learning Transferable Visual Models From Natural Language Supervision},
  author =       {Radford, Alec and Kim, Jong Wook and Hallacy, Chris and Ramesh, Aditya and Goh, Gabriel and Agarwal, Sandhini and Sastry, Girish and Askell, Amanda and Mishkin, Pamela and Clark, Jack and Krueger, Gretchen and Sutskever, Ilya},
  booktitle = 	 {Proceedings of the 38th International Conference on Machine Learning},
  pages = 	 {8748--8763},
  year = 	 {2021},
  editor = 	 {Meila, Marina and Zhang, Tong},
  volume = 	 {139},
  series = 	 {Proceedings of Machine Learning Research},
  month = 	 {18--24 Jul},
  publisher =    {PMLR}
}

@inproceedings{
bolya2025perception,
title={Perception Encoder: The best visual embeddings are not at the output of the network},
author={Daniel Bolya and Po-Yao Huang and Peize Sun and Jang Hyun Cho and Andrea Madotto and Chen Wei and Tengyu Ma and Jiale Zhi and Jathushan Rajasegaran and Hanoona Abdul Rasheed and Junke Wang and Marco Monteiro and Hu Xu and Shiyu Dong and Nikhila Ravi and Shang-Wen Li and Piotr Dollar and Christoph Feichtenhofer},
booktitle={The Thirty-ninth Annual Conference on Neural Information Processing Systems},
year={2025},
}

@inproceedings{guo2017calibration,
  title={On calibration of modern neural networks},
  author={Guo, Chuan and Pleiss, Geoff and Sun, Yu and Weinberger, Kilian Q},
  booktitle={International conference on machine learning},
  pages={1321--1330},
  year={2017},
  organization={PMLR}
}

@inproceedings{ismail2024conceptbottleneckgenerative,
  title={Concept bottleneck generative models},
  author={Ismail, Aya Abdelsalam and Adebayo, Julius and Corrada Bravo, Hector and Ra, Stephen and Cho, Kyunghyun},
  booktitle={International Conference on Learning Representations},
  volume={2024},
  pages={34114--34132},
  year={2024}
}

@inproceedings{martinezgarcia2026probabilistichardconcept,
  title={A probabilistic hard concept bottleneck for steerable generative models},
  author={Mart{\'\i}nez-Garc{\'\i}a, Mar{\'\i}a and Vazquez Alvarez, Ricardo and Lancho, Alejandro and Olmos, Pablo and Valera, Isabel},
  booktitle={International Conference on Learning Representations},
  volume={2026},
  pages={83758--83790},
  year={2026}
}

@article{labs2025flux1kontext,
  title={Flux. 1 kontext: Flow matching for in-context image generation and editing in latent space},
  author={Labs, Black Forest and Batifol, Stephen and Blattmann, Andreas and Boesel, Frederic and Consul, Saksham and Diagne, Cyril and Dockhorn, Tim and English, Jack and English, Zion and Esser, Patrick and others},
  journal={arXiv preprint arXiv:2506.15742},
  year={2025}
}

@inproceedings{
havasi2022addressing,
title={Addressing Leakage in Concept Bottleneck Models},
author={Marton Havasi and Sonali Parbhoo and Finale Doshi-Velez},
booktitle={Advances in Neural Information Processing Systems},
editor={Alice H. Oh and Alekh Agarwal and Danielle Belgrave and Kyunghyun Cho},
year={2022},
}
\bibliographystyle{iclr2027_conference}

\clearpage
\appendix
\addtocontents{toc}{\protect\setcounter{tocdepth}{2}}
\renewcommand{\contentsname}{Appendix contents}
{\small\tableofcontents}
\newpage
\section{Additional quantitative evaluation}
\label{app:additional-quantitative}

\subsection{Architecture ablations: graph layers and top-k}
\label{app:architecture-ablations}

TRACE uses one graph layer in the shared stack and one layer in each task-specific branch.
It retains $k{=}20$ spatial and cross-temporal messages per target concept, and $k{=}20$ same-concept temporal channels globally, following the spatio-temporal message-passing perspective of \citet{yan2018spatial}.
The activity-feedback graph retains $k{=}40$ label-to-concept messages.
These are the default settings for the proposed model's main and supplementary results.
Table~\ref{tab:repro-proposed-hparams} gives the complete topology, including the gradual top-$k$ warm-up schedule.
The transition-factorial structural ablation in the main paper separately removes recursive transitions, graph structure, activity feedback, and score normalization while retaining the rest of the training protocol.
We additionally ran in Table~\ref{tab:architecture-sensitivity} a sensitivity sweep on the three single-split main protocols.
The sweep varies graph depth, isolates each concept-relation family while keeping activity feedback and other settings fixed, and doubles all three concept-relation budgets.

\begin{table}[h]
    \centering
    \caption{\textbf{Graph architecture and topology sensitivity.} PE-L14, 128-concept results for BARISTA , Breakfast s1, and MPII Attr single-split protocols.
    Values are H3 top-1 accuracy in percent, reported as mean $\pm$ sample standard deviation over seeds 42-44.
    Depth rows jointly set the shared and task-specific graph-stack depth; one-hot rows retain only the named concept-relation family while keeping activity feedback at its default setting; the final row uses $k{=}40$ for all three concept-relation families. Best result per dataset is bold; second best is underlined.}
    \label{tab:architecture-sensitivity}
    \begingroup
    \scriptsize
    \setlength{\tabcolsep}{4pt}
    \begin{tabular}{lccc}
        \toprule
        Variant & BARISTA & Breakfast s1 & MPII Attr \\
        \midrule
        Full (1 layer, $k{=}20$) & $\mval{67.4}{0.3}$ & \underline{$\mval{42.9}{0.2}$} & $\mval{38.7}{0.6}$ \\
        2 graph layers & $\mval{68.0}{0.4}$ & $\mval{42.8}{1.1}$ & $\mval{38.5}{1.0}$ \\
        3 graph layers & \underline{$\mval{68.2}{0.3}$} & $\bmval{43.3}{0.6}$ & \underline{$\mval{38.8}{0.3}$} \\
        4 graph layers & $\bmval{68.5}{0.1}$ & \underline{$\mval{42.9}{1.3}$} & $\mval{38.0}{0.3}$ \\
        Spatial only & $\mval{67.5}{0.5}$ & $\mval{42.4}{0.7}$ & $\mval{38.5}{0.5}$ \\
        Same-concept temporal only & $\mval{67.3}{0.3}$ & $\mval{42.8}{0.6}$ & $\bmval{39.2}{0.3}$ \\
        Cross-temporal only & $\mval{67.4}{0.6}$ & \underline{$\mval{42.9}{0.2}$} & $\mval{38.4}{0.5}$ \\
        All families, $k{=}40$ & $\mval{67.8}{0.2}$ & $\mval{42.8}{0.4}$ & $\mval{38.5}{0.5}$ \\
        \bottomrule
    \end{tabular}%
    \endgroup
\end{table}

The depth-3 arm has the highest unweighted dataset-macro H3 top-1 accuracy, about 0.4 percentage points above the full setting.
Depth 4 improves BARISTA but reduces MPII, while one-hot relation effects are family- and dataset-dependent.
Doubling all concept-relation budgets to $k{=}40$ changes H3 top-1 by at most 0.5 percentage points.
The selected one-layer-per-stack, $k{=}20$ configuration is therefore a stable balanced default setting rather than an accuracy-optimal choice on every dataset.

\subsection{Matched transition-factorial ablation}
\label{app:transition-factorial}

Table~\ref{tab:transition-factorial-full} gives the complete activity and H3 statistics underlying Table~\ref{tab:component-ablation}.
The dense/no-graph arm is not parameter matched to the graph arm and is interpreted as a structural comparison.
Persistence copies the final observed concept state at every future horizon and disables graph processing and activity feedback.
Dense rollout uses a dense observed-state refiner and dense recursive forecast, with activity feedback disabled.
The graph-no-feedback arm retains graph refinement and recursive graph rollout but removes predicted-activity feedback, while the no-score-normalization arm disables the learned concept-score normalizer.
The full arm is graph rollout with activity feedback and score normalization, so it is not duplicated as a separate graph-with-feedback row.

\begin{table*}[t]
    \centering
    \caption{\textbf{Transition-factorial structural ablation.} PE-L14, 128-concept results on BARISTA, Breakfast s1, and MPII Attr protocols, reported as mean $\pm$ sample standard deviation over seeds 42--44 in percent. The dense/no-graph arm is not parameter matched to the graph arm. Best result per dataset and metric is bold; second best is underlined.}
    \label{tab:transition-factorial-full}
    \begingroup
    \scriptsize
    \setlength{\tabcolsep}{3pt}

        \begin{tabular}{llrrr}
        \toprule
        Dataset & Variant & Activity top-1 & H3 top-1 & H3 top-3 \\
        \midrule
        \multirow{5}{*}{BARISTA}
        & Full model & $\mval{76.6}{2.9}$ & \underline{$\mval{67.4}{0.3}$} & \underline{$\mval{81.8}{1.5}$} \\
        & Persistence & \underline{$\mval{77.1}{0.8}$} & $\mval{66.0}{0.2}$ & $\mval{79.5}{0.2}$ \\
        & Dense rollout, no graph & $\bmval{81.1}{0.8}$ & $\bmval{70.2}{0.3}$ & $\bmval{85.1}{0.6}$ \\
        & Graph, no activity feedback & $\mval{76.4}{1.8}$ & $\mval{67.0}{0.1}$ & $\mval{81.4}{0.9}$ \\
        & No score normalization & $\mval{75.3}{0.5}$ & $\mval{64.3}{0.5}$ & $\mval{78.5}{0.4}$ \\
        \midrule
        \multirow{5}{*}{Breakfast s1}
        & Full model & $\bmval{51.7}{0.4}$ & $\bmval{42.9}{0.2}$ & $\bmval{68.9}{0.7}$ \\
        & Persistence & $\mval{49.1}{0.2}$ & $\bmval{42.9}{0.1}$ & $\mval{66.2}{0.0}$ \\
        & Dense rollout, no graph & $\mval{50.5}{0.4}$ & $\mval{40.2}{0.3}$ & $\mval{66.5}{0.7}$ \\
        & Graph, no activity feedback & \underline{$\mval{50.9}{0.3}$} & \underline{$\mval{42.5}{0.6}$} & \underline{$\mval{67.1}{0.3}$} \\
        & No score normalization & $\mval{46.5}{0.8}$ & $\mval{39.6}{0.7}$ & $\mval{64.6}{0.8}$ \\
        \midrule
        \multirow{5}{*}{MPII Attr}
        & Full model & \underline{$\mval{47.0}{0.6}$} & $\bmval{38.7}{0.6}$ & \underline{$\mval{58.5}{0.4}$} \\
        & Persistence & $\mval{45.6}{0.2}$ & \underline{$\mval{38.2}{0.3}$} & $\bmval{59.2}{0.1}$ \\
        & Dense rollout, no graph & $\bmval{48.8}{0.4}$ & $\mval{35.9}{0.3}$ & $\mval{57.2}{1.0}$ \\
        & Graph, no activity feedback & $\mval{46.6}{0.1}$ & $\bmval{38.7}{0.4}$ & $\bmval{59.2}{0.5}$ \\
        & No score normalization & $\mval{43.0}{0.2}$ & $\mval{35.0}{0.3}$ & $\mval{55.4}{0.3}$ \\
        \bottomrule
    \end{tabular}%
    \endgroup
\end{table*}

The full configuration has the highest unweighted average H3 top-1 accuracy across the three protocols, but no single transition component improves every dataset.
In particular, score normalization is consistently important, whereas dense rollout helps BARISTA but harms the two cooking datasets and activity feedback has little short-horizon accuracy effect.
The graph should therefore be interpreted as an inspectable transition parameterization for the model-edit interface rather than as a universally accuracy-improving or causal structure.

\newpage
\subsection{Loss-objective ablation}
\label{app:loss-ablation}

Table~\ref{tab:loss-ablation} reports a loss-objective ablation each non-full arm removes one loss component: concept-forecast supervision, transition tolerance, the two SIL false-positive penalties, or classifier $L_1$ regularization.
We designed the ablation as a component diagnostic rather than a search for a dataset-specific accuracy-maximizing objective: each term supports a different modeling requirement, and removing one changes the preferred operating point rather than uniformly improving the model. 
The full objective is the fixed configuration used in the main experiments.
Each term can be turned off for diagnosis, but the resulting arms optimize different and sometimes competing properties: some removals improve selected accuracy values, whereas the SIL penalty reduces false positives and classifier $L_1$ preserves selective named-concept contributions (Tables~\ref{tab:loss-ablation-selectivity} and~\ref{tab:loss-ablation-node-intervention}). 

\begin{table*}[h]
    \centering
    \caption{\textbf{Loss-objective ablation.} PE-L14, 128-concept results on the BARISTA split, Breakfast s1, and MPII Attr, reported as mean $\pm$ sample standard deviation over seeds 42--44. Each non-full arm removes one loss component while keeping the architecture and training protocol fixed. Metrics are activity top-1 (Act.), H3 top-1 (H3-1), and H3 top-3 (H3-3) accuracy in percent. Best result per dataset and metric is bold.}
    \label{tab:loss-ablation}
    \begingroup
    \scriptsize
    \setlength{\tabcolsep}{3pt}
    \resizebox{\textwidth}{!}{%
    \begin{tabular}{lccccccccc}
        \toprule
        & \multicolumn{3}{c}{BARISTA} & \multicolumn{3}{c}{Breakfast s1} & \multicolumn{3}{c}{MPII Attr} \\
        \cmidrule(lr){2-4} \cmidrule(lr){5-7} \cmidrule(lr){8-10}
        Loss variant & Act. & H3-1 & H3-3 & Act. & H3-1 & H3-3 & Act. & H3-1 & H3-3 \\
        \midrule
        Full loss
        & $\mval{76.6}{2.9}$ & $\mval{67.4}{0.3}$ & $\mval{81.8}{1.5}$
        & $\mval{51.7}{0.4}$ & $\bmval{42.9}{0.2}$ & $\mval{68.9}{0.7}$
        & $\mval{47.0}{0.6}$ & $\mval{38.7}{0.6}$ & $\mval{58.5}{0.4}$ \\
        No concept-forecast loss
        & $\mval{78.0}{2.4}$ & $\mval{67.6}{0.4}$ & $\mval{82.3}{1.4}$
        & $\mval{51.7}{0.4}$ & $\mval{42.8}{0.4}$ & $\bmval{69.0}{0.5}$
        & $\mval{46.9}{0.5}$ & $\mval{38.6}{0.7}$ & $\mval{58.6}{0.5}$ \\
        No transition tolerance
        & $\mval{78.0}{2.7}$ & $\mval{67.3}{0.5}$ & $\mval{82.5}{1.5}$
        & $\mval{51.7}{0.4}$ & $\bmval{42.9}{0.3}$ & $\bmval{69.0}{0.6}$
        & $\mval{47.2}{0.4}$ & $\mval{38.6}{0.7}$ & $\mval{58.4}{0.7}$ \\
        No SIL penalty
        & $\mval{77.9}{0.6}$ & $\bmval{68.5}{0.2}$ & $\bmval{84.3}{0.4}$
        & $\mval{51.5}{0.4}$ & $\mval{42.7}{0.5}$ & $\mval{68.8}{0.6}$
        & $\mval{47.0}{0.5}$ & $\mval{38.3}{0.7}$ & $\bmval{59.2}{0.5}$ \\
        No classifier $L_1$
        & $\bmval{80.3}{0.3}$ & $\mval{67.7}{0.5}$ & $\mval{83.9}{0.2}$
        & $\bmval{51.9}{0.1}$ & $\mval{42.9}{0.1}$ & $\mval{68.2}{0.8}$
        & $\bmval{48.1}{0.5}$ & $\bmval{39.1}{0.5}$ & $\mval{58.8}{0.3}$ \\
        \bottomrule
    \end{tabular}%
    }
    \endgroup
\end{table*}

Table~\ref{tab:loss-ablation-selectivity} reports diagnostics targeted to the intended loss behaviors.
The SIL false-positive rate measures background calibration, while predictor effective support and top-5 mass measure how selectively the shared activity predictor uses named concepts.
For a predictor row $w$, effective support is $\lVert w\rVert_1^2/\lVert w\rVert_2^2$, lower support and higher top-5 mass indicate a more concentrated, inspectable predictor.

\begin{table*}[h]
    \centering
    \caption{\textbf{SIL calibration and predictor selectivity in the loss ablation.} Each cell reports H3 SIL false-positive rate $\downarrow$ / predictor effective support $\downarrow$ / predictor top-5 mass $\uparrow$. Values are mean $\pm$ sample standard deviation over seeds 42--44 from the matched PE-L14, 128-concept checkpoints. SIL-FP is a percentage over true non-SIL targets; effective support is a concept count; top-5 mass is a percentage of absolute predictor weight.}
    \label{tab:loss-ablation-selectivity}
    \begingroup
    \scriptsize
    \setlength{\tabcolsep}{3pt}
    \resizebox{\textwidth}{!}{%
    \begin{tabular}{lccc}
        \toprule
        Loss variant & BARISTA (SIL-FP / $N_{\mathrm{eff}}$ / Top-5) & Breakfast s1 (SIL-FP / $N_{\mathrm{eff}}$ / Top-5) & MPII Attr (SIL-FP / $N_{\mathrm{eff}}$ / Top-5) \\
        \midrule
        Full loss
        & $\mval{39.1}{5.4}$ / $\mval{31.8}{10.1}$ / $\mval{37.8}{15.8}$
        & $\mval{7.1}{2.2}$ / $\mval{42.6}{0.7}$ / $\mval{27.7}{0.5}$
        & $\mval{12.6}{1.3}$ / $\mval{30.5}{0.5}$ / $\mval{40.8}{0.3}$ \\
        No concept-forecast loss
        & $\mval{36.0}{4.5}$ / $\mval{26.3}{9.3}$ / $\mval{46.6}{14.5}$
        & $\mval{7.1}{2.4}$ / $\mval{42.8}{0.7}$ / $\mval{27.6}{0.6}$
        & $\mval{12.1}{1.2}$ / $\mval{30.8}{0.6}$ / $\mval{40.6}{0.4}$ \\
        No transition tolerance
        & $\mval{36.0}{5.8}$ / $\mval{25.9}{10.3}$ / $\mval{46.2}{14.6}$
        & $\mval{7.0}{2.3}$ / $\mval{42.7}{0.6}$ / $\mval{27.6}{0.5}$
        & $\mval{12.6}{1.2}$ / $\mval{30.3}{1.0}$ / $\mval{40.9}{0.9}$ \\
        No SIL penalty
        & $\mval{43.5}{0.9}$ / $\mval{20.6}{1.7}$ / $\bmval{56.0}{1.2}$
        & $\mval{9.9}{2.8}$ / $\mval{42.5}{0.6}$ / $\mval{27.8}{0.5}$
        & $\mval{23.0}{2.1}$ / $\mval{30.6}{0.5}$ / $\mval{40.8}{0.5}$ \\
        No classifier $L_1$
        & $\mval{31.0}{0.8}$ / $\mval{82.4}{0.3}$ / $\mval{11.6}{0.0}$
        & $\mval{7.3}{2.3}$ / $\mval{83.2}{0.5}$ / $\mval{11.4}{0.1}$
        & $\mval{13.4}{0.9}$ / $\mval{81.1}{0.8}$ / $\mval{12.0}{0.2}$ \\
        \bottomrule
    \end{tabular}%
    }
    \endgroup
\end{table*}

Removing classifier $L_1$ directly makes the predictor less selective: effective support expands from $31.8$ to $82.4$ concepts on BARISTA, from $42.6$ to $83.2$ on Breakfast, and from $30.5$ to $81.1$ on MPII, while top-5 mass falls to $11.4$--$12.0\%$.
Without this penalty, each activity prediction depends on roughly 81--83 concepts, making its contribution pattern difficult to inspect. 
The classifier penalty is therefore needed to keep named-concept explanations selective.
Removing the SIL penalty raises H3 SIL false positives from $39.1\%$ to $43.5\%$ on BARISTA, from $7.1\%$ to $9.9\%$ on Breakfast, and from $12.6\%$ to $23.0\%$ on MPII.

Table~\ref{tab:loss-ablation-node-intervention} evaluates the same loss arms with matched oracle-ranked concept-node edits.
The 30 initially wrong and 30 initially correct video windows are selected once from the full-loss checkpoint for each dataset and seed, then reused for every loss arm.
This isolates loss effects from changes in the sampled cases.
The first value in each cell is the wrong-to-correct rate: for the full-loss arm it is $25.6\%$, $25.6\%$, and $34.4\%$ on BARISTA, Breakfast, and MPII, respectively, matching the concept-node results in Table~\ref{tab:intervention-interfaces}.
The bold $0.0$ values are the second entry, correct-to-wrong rate, and indicate that these edits caused no observed harm rather than no correction.

\begin{table*}[h]
    \centering
    \caption{\textbf{Matched oracle node edits in the loss ablation.} Budget-5 directional-oracle concept-node edits on the same 30 wrong and 30 correct video windows per dataset and seed selected from the full-loss checkpoint. Entries report wrong-to-correct rate $\uparrow$ / correct-to-wrong rate $\downarrow$ / true-class probability change $\uparrow$ in percentage points, mean $\pm$ sample standard deviation over seeds 42--44. The future concept target is privileged, so these are oracle upper bounds. }
    \label{tab:loss-ablation-node-intervention}
    \begingroup
    \scriptsize
    \setlength{\tabcolsep}{3pt}
    \resizebox{\textwidth}{!}{%
    \begin{tabular}{lccc}
        \toprule
        Loss variant & BARISTA (W$\rightarrow$C / C$\rightarrow$W / $\Delta p$) & Breakfast s1 (W$\rightarrow$C / C$\rightarrow$W / $\Delta p$) & MPII Attr (W$\rightarrow$C / C$\rightarrow$W / $\Delta p$) \\
        \midrule
        Full loss
        & $\mval{25.6}{18.4}$ / $\mval{0.0}{0.0}$ / $\mval{7.17}{5.26}$
        & $\mval{25.6}{3.8}$ / $\mval{0.0}{0.0}$ / $\mval{11.35}{1.20}$
        & $\mval{34.4}{6.9}$ / $\mval{0.0}{0.0}$ / $\mval{9.84}{1.53}$ \\
        No concept-forecast loss
        & $\mval{29.9}{22.0}$ / $\mval{0.0}{0.0}$ / $\mval{9.33}{4.54}$
        & $\mval{23.6}{3.0}$ / $\mval{0.0}{0.0}$ / $\mval{11.89}{1.10}$
        & $\mval{38.4}{6.9}$ / $\mval{0.0}{0.0}$ / $\mval{11.71}{1.51}$ \\
        No transition tolerance
        & $\mval{31.1}{15.9}$ / $\mval{0.0}{0.0}$ / $\mval{9.10}{3.93}$
        & $\mval{26.9}{5.4}$ / $\mval{0.0}{0.0}$ / $\mval{11.43}{1.30}$
        & $\mval{36.3}{7.2}$ / $\mval{0.0}{0.0}$ / $\mval{9.81}{1.46}$ \\
        No SIL penalty
        & $\mval{40.7}{12.9}$ / $\mval{0.0}{0.0}$ / $\mval{10.61}{2.91}$
        & $\mval{25.8}{1.7}$ / $\mval{0.0}{0.0}$ / $\mval{11.56}{1.19}$
        & $\mval{34.1}{7.1}$ / $\mval{0.0}{0.0}$ / $\mval{10.25}{1.27}$ \\
        No classifier $L_1$
        & $\mval{42.2}{13.2}$ / $\mval{0.0}{0.0}$ / $\mval{12.06}{1.46}$
        & $\mval{33.0}{6.7}$ / $\mval{1.3}{2.3}$ / $\mval{13.71}{1.84}$
        & $\mval{42.1}{11.7}$ / $\mval{1.1}{2.0}$ / $\mval{11.26}{0.77}$ \\
        \bottomrule
    \end{tabular}%
    }
    \endgroup
\end{table*}

The oracle node edits improve initially wrong forecasts for every loss arm, but removing classifier $L_1$ produces the largest gains and also introduces small correct-to-wrong rates on Breakfast and MPII.
This is consistent with the selectivity result: a less concentrated predictor is more sensitive to edits, but its edits are less controlled.
The SIL and predictor diagnostics therefore expose loss effects that endpoint accuracy alone does not reveal.

\subsection{Matched non-graph dynamics control}
\label{app:matched-non-graph-dynamics}

Table~\ref{tab:matched-non-graph-dynamics} reports the retrained graph-free control with exactly the same trainable parameter count as the graph model for each dataset.
The control replaces both observed-state and recursive forecast graph layers with graph-layer-matched dense residual layers, while retaining the same concept vocabulary, losses, splits, history, horizon, and seeds.
The existing dense/no-graph arm in Table~\ref{tab:transition-factorial-full} is not parameter matched and should not be used for this comparison.

\begin{table}[h]
    \centering
    \caption{\textbf{Exact-capacity graph-free dynamics control.} PE-L14, 128-concept results on BARISTA, Breakfast s1, and MPII Attr protocols, reported over seeds 42--44. The graph-free control uses dense residual transitions with exactly the graph model's trainable parameter count; all other protocol settings are held fixed.}
    \label{tab:matched-non-graph-dynamics}
    \begingroup
    \scriptsize
    \setlength{\tabcolsep}{3pt}
    \begin{tabular}{llrrrr}
        \toprule
        Dataset & Dynamics & Trainable params & Activity top-1 & H3 top-1 & H3 top-3 \\
        \midrule
        \multirow{2}{*}{BARISTA}
        & Graph & 295,153 & $\mval{76.56}{2.89}$ & $\mval{67.36}{0.30}$ & $\mval{81.78}{1.52}$ \\
        & Exact-capacity graph-free & 295,153 & $\mval{81.08}{0.83}$ & $\mval{70.37}{0.38}$ & $\mval{85.27}{0.60}$ \\
        \midrule
        \multirow{2}{*}{Breakfast}
        & Graph & 271,668 & $\mval{51.70}{0.35}$ & $\mval{42.89}{0.23}$ & $\mval{68.89}{0.73}$ \\
        & Exact-capacity graph-free & 271,668 & $\mval{50.64}{0.80}$ & $\mval{41.29}{0.92}$ & $\mval{66.75}{2.61}$ \\
        \midrule
        \multirow{2}{*}{MPII Attr}
        & Graph & 286,683 & $\mval{47.03}{0.58}$ & $\mval{38.65}{0.64}$ & $\mval{58.54}{0.39}$ \\
        & Exact-capacity graph-free & 286,683 & $\mval{48.27}{0.47}$ & $\mval{38.05}{0.79}$ & $\mval{58.01}{1.23}$ \\
        \bottomrule
    \end{tabular}%
    \endgroup
\end{table}

Exact parameter matching does not produce a uniform forecasting advantage for either transition mechanism. 
The graph-free control is higher on BARISTA, whereas the graph is higher on Breakfast. 
On MPII Attr, the graph-free control is higher on activity top-1, while the graph is higher on H3 top-1 and top-3. 
As discussed above, the graph does not automatically improve predictive performance, but it provides an explicit path for concept propagation.

\subsection{Forecasting and probability diagnostics}
\label{app:forecast-probability}

Table~\ref{tab:forecast-probability} reports the full-model H1--H3 diagnostics.
NLL measures the negative log probability assigned to the target class, multiclass Brier measures squared probability-vector error, and 15-bin ECE measures the discrepancy between confidence and empirical correctness~\citep{guo2017calibration}, so lower values are better.
The metrics are computed over every valid H3 evaluation window.

\begin{table}[h]
    \centering
    \caption{\textbf{Full-model forecasting and probability diagnostics.} PE-L14, 128-concept results on BARISTA, Breakfast s1, and MPII Attr protocols for H1--H3, reported over seeds 42--44. NLL is negative log-likelihood, Brier is the multiclass Brier score, and ECE is fixed-width 15-bin expected calibration error.}
    \label{tab:forecast-probability}
    \begingroup
    \scriptsize
    \setlength{\tabcolsep}{3pt}
    \begin{tabular}{llccccc}
        \toprule
        Dataset & Horizon & Top-1 & Top-3 & NLL $\downarrow$ & Brier $\downarrow$ & ECE $\downarrow$ \\
        \midrule
        \multirow{3}{*}{BARISTA}
        & H1 & $\mval{74.8}{1.1}$ & $\mval{87.2}{1.3}$ & $\mval{1.054}{0.109}$ & $\mval{0.357}{0.026}$ & $\mval{0.093}{0.044}$ \\
        & H2 & $\mval{69.5}{0.9}$ & $\mval{83.6}{1.3}$ & $\mval{1.174}{0.071}$ & $\mval{0.406}{0.012}$ & $\mval{0.039}{0.006}$ \\
        & H3 & $\mval{67.4}{0.3}$ & $\mval{81.8}{1.5}$ & $\mval{1.284}{0.059}$ & $\mval{0.444}{0.009}$ & $\mval{0.064}{0.002}$ \\
        \midrule
        \multirow{3}{*}{Breakfast s1}
        & H1 & $\mval{48.8}{0.2}$ & $\mval{73.9}{0.6}$ & $\mval{1.768}{0.015}$ & $\mval{0.669}{0.003}$ & $\mval{0.058}{0.006}$ \\
        & H2 & $\mval{45.8}{0.3}$ & $\mval{71.5}{0.7}$ & $\mval{1.981}{0.021}$ & $\mval{0.734}{0.005}$ & $\mval{0.157}{0.005}$ \\
        & H3 & $\mval{42.9}{0.2}$ & $\mval{68.9}{0.7}$ & $\mval{2.303}{0.028}$ & $\mval{0.811}{0.005}$ & $\mval{0.251}{0.004}$ \\
        \midrule
        \multirow{3}{*}{MPII Attr}
        & H1 & $\mval{44.8}{0.5}$ & $\mval{65.2}{0.2}$ & $\mval{2.080}{0.010}$ & $\mval{0.702}{0.004}$ & $\mval{0.032}{0.010}$ \\
        & H2 & $\mval{41.7}{0.5}$ & $\mval{62.0}{0.5}$ & $\mval{2.285}{0.020}$ & $\mval{0.756}{0.006}$ & $\mval{0.100}{0.011}$ \\
        & H3 & $\mval{38.7}{0.6}$ & $\mval{58.5}{0.4}$ & $\mval{2.558}{0.032}$ & $\mval{0.819}{0.009}$ & $\mval{0.170}{0.012}$ \\
        \bottomrule
    \end{tabular}%
    \endgroup
\end{table}

Table~\ref{tab:main-sil-diagnostics} isolates the effect of unlabeled windows.
Rates are computed over valid H3 sliding-window test examples, non-SIL accuracy is conditioned on true non-SIL targets, and non-SIL macro-F1 averages only the non-SIL activity classes while treating SIL predictions as errors.

\begin{table}[h]
    \centering
    \caption{\textbf{Main-dataset H3 SIL diagnostics.} Results use 128 concepts, PE-L14, the BARISTA split, Breakfast s1, and MPII Attr, with seeds 42--44. Values are mean $\pm$ sample standard deviation across seeds; $N$ is the number of valid H3 windows per seed and $N_{\mathrm{non\text{-}SIL}}$ is the number whose ground-truth target is non-SIL.}
    \label{tab:main-sil-diagnostics}
    \begingroup
    \scriptsize
    \setlength{\tabcolsep}{4pt}
    \begin{tabular}{lccccc}
        \toprule
        Dataset & $N$ & $N_{\mathrm{non\text{-}SIL}}$ & True SIL $\%$ & Predicted SIL $\%$ & Non-SIL acc. / macro-F1 $\%$ \\
        \midrule
        BARISTA & 2,415 & 853 & $64.68 $ & $\mval{72.17}{2.57}$ & $\mval{25.48}{2.72}$ / $\mval{5.65}{0.73}$ \\
        Breakfast & 16,770 & 15,780 & $5.90 $ & $\mval{7.69}{2.30}$ & $\mval{44.50}{0.43}$ / $\mval{23.66}{0.27}$ \\
        MPII Attr & 12,956 & 10,425 & $19.54 $ & $\mval{19.56}{1.39}$ & $\mval{36.31}{1.22}$ / $\mval{11.08}{0.30}$ \\
        \bottomrule
    \end{tabular}%
    \endgroup
\end{table}

Table~\ref{tab:last-label-baseline} provides a direct horizon-wise last-label reference for the forecasting claims.
It copies the activity label at the end of the observed history to every future horizon, unlike the learned concept-persistence arm in the transition factorial, it has no trainable parameters.
As the table illustrates, activities often persist across consecutive horizons, which explains why simply repeating the last observed label is such a strong baseline.
Although TRACE underperforms this baseline at horizons H1 and H2, this does not mean that its forecasts are inferior. 
Rather, TRACE learns to predict transitions to different activity classes instead of merely preserving the previous label.

\begin{table}[h]
    \centering
    \caption{\textbf{Matched last-label forecasting baseline.} Top-1 accuracy for H1--H3 on the same 128-concept PE-L14 protocols and valid windows as the forecasting diagnostics, using seeds 42--44. The baseline repeats the final observed activity label; because the targets and windows are identical across seeds, the seed-level standard deviation is zero.}
    \label{tab:last-label-baseline}
    \begingroup
    \scriptsize
    \setlength{\tabcolsep}{5pt}
    \begin{tabular}{lcccc}
        \toprule
        Dataset & $N$ per horizon & H1 top-1 $\%$ & H2 top-1 $\%$ & H3 top-1 $\%$ \\
        \midrule
        BARISTA & 2,415 & $80.66 $ & $\mval{67.87}{0.00}$ & $\mval{62.36}{0.00}$ \\
        Breakfast & 16,770 & $91.12 $ & $\mval{82.61}{0.00}$ & $\mval{75.01}{0.00}$ \\
        MPII Attr & 12,956 & $81.24 $ & $\mval{66.89}{0.00}$ & $\mval{57.25}{0.00}$ \\
        \bottomrule
    \end{tabular}%
    \endgroup
\end{table}

\subsection{Forecasting on activity changes and non-SIL targets}
\label{app:change-forecast}

Table~\ref{tab:change-forecast} evaluates all H3 windows where the label changes and all non-SIL targets (five seeds, identical windows for all models). Black-box models are stronger on changed windows on BARISTA and MPII, TRACE has the best non-SIL macro-F1 on Breakfast and MPII.

\begin{table}[h!]
    \centering
    \caption{\textbf{Forecasting on activity changes and non-SIL targets.} H3 top-1 on windows with a label change / non-SIL top-1 / non-SIL macro-F1 (\%), five seeds.}
    \label{tab:change-forecast}
    \scriptsize
    \begin{tabular}{lccc}
        \toprule
        Model & BARISTA & Breakfast s1 & MPII Attr \\
        \midrule
        TRACE & 21.0 / 25.0 / 5.5 & 16.1 / 44.3 / 23.7 & 16.3 / 36.4 / 11.1 \\
        Sparse Linear & 23.7 / 27.4 / 5.7 & 15.2 / 43.9 / 20.7 & 16.9 / 35.8 / 9.6 \\
        Feat. Transf. & 29.1 / 32.6 / 5.8 & 17.5 / 46.0 / 18.8 & 18.6 / 38.0 / 9.8 \\
        SlowFast & 30.7 / 32.1 / 7.1 & 18.2 / 45.6 / 19.0 & 19.3 / 37.7 / 10.1 \\
        \bottomrule
    \end{tabular}
\end{table}

\subsection{Generic shared concept vocabulary}
\label{app:generic-vocab}

We replaced the dataset-specific lists by one generic 128-concept vocabulary shared by all datasets (no activity names) and retrained TRACE (seeds 42--44). Accuracy stays within 2.5 points and oracle correction within seed noise (Table~\ref{tab:generic-vocab}). 
We keep specific lists as default, since every coordinate is something a user reads and edits.

\begin{table}[h]
    \centering
    \caption{\textbf{Specific vs.\ generic vocabulary.} Activity top-1 / H3 top-1 / final W$\rightarrow$C (\%, five oracle node edits); C$\rightarrow$W is 0 throughout.}
    \label{tab:generic-vocab}
    \scriptsize
    \begin{tabular}{lccc}
        \toprule
        Vocabulary & BARISTA & Breakfast s1 & MPII Attr \\
        \midrule
        Dataset-specific & 76.6 / 67.4 / 25.6 & 51.7 / 42.9 / 25.6 & 47.0 / 38.7 / 34.4 \\
        Generic shared & 74.2 / 67.3 / 17.8 & 51.0 / 42.4 / 34.4 & 49.5 / 40.0 / 37.8 \\
        \bottomrule
    \end{tabular}
\end{table}

\subsection{Further backbone ablations}
\label{app:pe-g14-main-three}

Table~\ref{tab:pe-g14-main-three} reports the PE-G14 result matrix, where PE-G14 and PE-L14 are variants of the Perception Encoder visual representation~\citep{bolya2025perception}.

\begin{table}[h]
    \centering
    \caption{\textbf{PE-G14 128-concept result matrix.} PE-G14 results on the BARISTA split, Breakfast s1--s4, and MPII Attr/Dishes, with activity classification and H3 forecast accuracy reported over seeds 42--46. Multi-split datasets are averaged within seed. Best result per dataset and metric is bold; second best is underlined.}
    \label{tab:pe-g14-main-three}
    \begingroup
    \scriptsize
    \setlength{\tabcolsep}{2pt}
    \resizebox{\textwidth}{!}{%
    \begin{tabular}{llcccccccccccc}
        \toprule
        & & \multicolumn{4}{c}{BARISTA} & \multicolumn{4}{c}{Breakfast} & \multicolumn{4}{c}{MPII Cooking~2} \\
        \cmidrule(lr){3-6} \cmidrule(lr){7-10} \cmidrule(lr){11-14}
        & & \multicolumn{2}{c}{Activity} & \multicolumn{2}{c}{H3} & \multicolumn{2}{c}{Activity} & \multicolumn{2}{c}{H3} & \multicolumn{2}{c}{Activity} & \multicolumn{2}{c}{H3} \\
        \cmidrule(lr){3-4} \cmidrule(lr){5-6} \cmidrule(lr){7-8} \cmidrule(lr){9-10} \cmidrule(lr){11-12} \cmidrule(lr){13-14}
        Group & Method & Top-1 & Top-3 & Top-1 & Top-3 & Top-1 & Top-3 & Top-1 & Top-3 & Top-1 & Top-3 & Top-1 & Top-3 \\
        \midrule
        Linear
        & Sparse Linear
        & $\mval{68.6}{0.7}$ & $\mval{79.0}{0.4}$ & $\mval{68.8}{0.3}$ & $\mval{82.9}{0.6}$
        & $\mval{46.4}{0.4}$ & $\mval{70.9}{0.4}$ & $\mval{39.6}{0.3}$ & $\mval{64.2}{0.4}$
        & $\mval{47.3}{0.5}$ & $\mval{66.2}{0.7}$ & $\mval{41.1}{0.7}$ & $\mval{62.2}{0.2}$ \\
        \midrule
        \multirow{2}{*}{Concept-based}
        & MoTIF
        & $\mval{73.7}{2.4}$ & $\mval{87.1}{1.1}$ & n/a & n/a
        & $\mval{47.8}{0.4}$ & $\mval{72.1}{0.4}$ & n/a & n/a
        & $\mval{49.6}{1.4}$ & $\mval{69.8}{0.6}$ & n/a & n/a \\
        & TRACE (ours)
        & $\mval{76.2}{0.5}$ & $\mval{89.3}{0.3}$ & $\mval{68.0}{0.1}$ & $\mval{82.1}{0.5}$
        & $\mval{48.7}{0.3}$ & $\bmval{73.4}{0.1}$ & $\mval{40.0}{0.3}$ & $\mval{64.7}{0.3}$
        & $\mval{51.4}{0.5}$ & \underline{$\mval{71.4}{0.2}$} & \underline{$\mval{42.0}{0.6}$} & $\mval{61.3}{0.3}$ \\
        \midrule
        \multirow{2}{*}{Black-box}
        & Feature Transformer
        & $\bmval{82.0}{0.4}$ & $\bmval{91.5}{0.2}$ & $\bmval{71.3}{0.3}$ & $\bmval{84.3}{0.2}$
        & $\bmval{50.6}{0.4}$ & \underline{$\mval{72.8}{0.4}$} & $\bmval{42.3}{0.4}$ & $\bmval{67.0}{0.3}$
        & $\bmval{52.2}{0.3}$ & $\bmval{71.7}{0.3}$ & $\bmval{42.4}{0.4}$ & \underline{$\mval{62.4}{0.4}$} \\
        & SlowFast-style
        & \underline{$\mval{81.9}{0.5}$} & \underline{$\mval{91.3}{0.4}$} & \underline{$\mval{69.6}{0.5}$} & \underline{$\mval{83.5}{0.4}$}
        & \underline{$\mval{49.4}{0.4}$} & $\mval{72.2}{0.6}$ & \underline{$\mval{41.6}{0.3}$} & \underline{$\mval{66.9}{0.2}$}
        & \underline{$\mval{51.2}{0.4}$} & $\mval{71.1}{0.6}$ & $\mval{41.9}{0.5}$ & $\bmval{62.8}{0.5}$ \\
        \bottomrule
    \end{tabular}%
    }
    \endgroup
\end{table}

Within this PE-G14 matrix, opaque feature models are strongest on aggregate accuracy, while TRACE remains competitive with the interpretable alternatives.
The Feature Transformer is best for H3 top-1 on all three datasets, and SlowFast-style is best for H3 top-3 on Breakfast and MPII Cooking~2, the two black-box models are nearly tied on BARISTA activity accuracy.
Among concept-based models, TRACE improves activity top-1 over MoTIF on BARISTA and MPII Cooking~2 and provides the comparable H3 concept rollout that MoTIF does not report.

\subsection{Concept ablations: 128, 256}
\label{sec:appendix-concept-budget-256}

Table~\ref{tab:matched-concept-budget} isolates TRACE and compares the two concept budgets.
This comparison tests concept budget, not independently generated vocabularies.

\begin{table}[h]
    \centering
    \caption{\textbf{Matched proposed-model concept-budget comparison.} PE-L14 proposed-model results for 128 versus 256 concepts on BARISTA, Breakfast s1--s4, and MPII Attr/Dishes, reported as mean $\pm$ sample standard deviation over seeds 42--46. Multi-split datasets are averaged within seed.}
    \label{tab:matched-concept-budget}
    \begingroup
    \scriptsize
    \setlength{\tabcolsep}{4pt}
    \begin{tabular}{lrrrr}
        \toprule
        Dataset & 128 activity & 256 activity & 128 H3 & 256 H3 \\
        \midrule
        BARISTA & $\mval{76.02}{2.18}$ & $\mval{79.71}{0.42}$ & $\mval{67.55}{0.35}$ & $\mval{68.51}{0.33}$ \\
        Breakfast & $\mval{45.52}{0.23}$ & $\mval{46.36}{0.21}$ & $\mval{37.48}{0.08}$ & $\mval{38.91}{0.19}$ \\
        MPII Attr/Dishes & $\mval{46.15}{0.43}$ & $\mval{47.68}{0.41}$ & $\mval{37.88}{0.46}$ & $\mval{39.40}{0.30}$ \\
        \bottomrule
    \end{tabular}%
    \endgroup
\end{table}

\begin{table}[h]
    \centering
    \caption{\textbf{PE-L14 256-concept result matrix.} PE-L14 results with 256 concepts on the BARISTA split, Breakfast s1--s4, and MPII Attr/Dishes, with activity classification and H3 forecast accuracy reported over seeds 42--46.
    Best result per dataset and metric is bold; second best is underlined.}
    \label{tab:blackbox-concept-baselines-256}
    \begingroup
    \scriptsize
    \setlength{\tabcolsep}{2pt}
    \resizebox{\textwidth}{!}{%
    \begin{tabular}{llcccccccccccc}
        \toprule
        & & \multicolumn{4}{c}{BARISTA} & \multicolumn{4}{c}{Breakfast} & \multicolumn{4}{c}{MPII Cooking~2} \\
        \cmidrule(lr){3-6} \cmidrule(lr){7-10} \cmidrule(lr){11-14}
        & & \multicolumn{2}{c}{Activity} & \multicolumn{2}{c}{H3} & \multicolumn{2}{c}{Activity} & \multicolumn{2}{c}{H3} & \multicolumn{2}{c}{Activity} & \multicolumn{2}{c}{H3} \\
        \cmidrule(lr){3-4} \cmidrule(lr){5-6} \cmidrule(lr){7-8} \cmidrule(lr){9-10} \cmidrule(lr){11-12} \cmidrule(lr){13-14}
        Group & Method & Top-1 & Top-3 & Top-1 & Top-3 & Top-1 & Top-3 & Top-1 & Top-3 & Top-1 & Top-3 & Top-1 & Top-3 \\
        \midrule
        Linear
        & Sparse Linear
        & $\mval{67.4}{0.6}$ & $\mval{79.7}{0.3}$ & \underline{$\mval{68.3}{0.1}$} & $\mval{82.4}{0.3}$
        & $\mval{43.7}{0.1}$ & $\mval{68.3}{0.1}$ & $\mval{37.7}{0.2}$ & $\mval{62.0}{0.2}$
        & $\mval{43.9}{0.5}$ & $\mval{61.3}{0.9}$ & $\mval{38.3}{0.5}$ & $\mval{58.8}{0.7}$ \\
        \midrule
        \multirow{2}{*}{Concept-based}
        & MoTIF
        & $\mval{75.7}{4.4}$ & $\mval{87.7}{2.1}$ & n/a & n/a
        & $\mval{45.7}{0.5}$ & $\mval{70.4}{0.5}$ & n/a & n/a
        & $\mval{45.6}{1.0}$ & $\mval{66.7}{0.7}$ & n/a & n/a \\
        & TRACE (ours)
        & $\bmval{79.7}{0.4}$ & \underline{$\mval{90.2}{0.2}$} & \underline{$\mval{68.5}{0.3}$} & $\bmval{83.6}{0.2}$
        & $\mval{46.4}{0.2}$ & $\bmval{71.2}{0.2}$ & $\mval{38.9}{0.2}$ & \underline{$\mval{63.9}{0.3}$}
        & $\mval{47.7}{0.4}$ & \underline{$\mval{67.2}{0.3}$} & $\mval{39.4}{0.3}$ & $\mval{59.5}{0.7}$ \\
        \midrule
        \multirow{2}{*}{Black-box}
        & Feature Transformer
        & $\mval{79.2}{0.3}$ & \underline{$\mval{89.3}{0.5}$} & $\bmval{69.6}{0.3}$ & \underline{$\mval{82.9}{0.3}$}
        & $\bmval{47.3}{0.4}$ & \underline{$\mval{69.6}{0.6}$} & \underline{$\mval{39.7}{0.6}$} & $\mval{63.7}{0.2}$
        & $\bmval{49.1}{1.0}$ & $\bmval{69.6}{0.7}$ & $\bmval{40.3}{0.9}$ & \underline{$\mval{61.0}{0.5}$} \\
        & SlowFast-style
        & \underline{$\mval{79.6}{0.5}$} & $\bmval{90.3}{0.9}$ & $\mval{67.6}{1.0}$ & $\mval{82.5}{0.8}$
        & \underline{$\mval{46.8}{0.5}$} & $\mval{69.5}{0.6}$ & $\bmval{39.7}{0.6}$ & $\bmval{64.1}{0.4}$
        & \underline{$\mval{48.1}{0.8}$} & \underline{$\mval{69.2}{0.7}$} & \underline{$\mval{39.6}{0.4}$} & $\bmval{61.0}{0.4}$ \\
        \bottomrule
    \end{tabular}%
    }
    \endgroup
\end{table}

Table~\ref{tab:blackbox-concept-baselines-256} reports the 256-concept result matrix.

The 256-concept results provide an accuracy-oriented operating point, with especially strong BARISTA activity recognition and Breakfast/MPII H3 forecasting.

\subsection{Additional datasets: GTEA and EPIC-KITCHENS-100}
\label{app:gtea_epic}
Table~\ref{tab:gtea-epic-128-concept-diagnostics} reports results for the two diagnostic stress-test datasets.
The results confirm that aggregate accuracy alone is misleading on these datasets.
On GTEA, the models predict SIL for approximately $86$--$98\%$ of H3 windows despite a true rate of $49.41\%$.
On EPIC-KITCHENS-100, the black-box baselines predict SIL for approximately $67$--$74\%$ of H3 windows, with $74\%$ for Sparse Linear and $67\%$ for the proposed model; all remain far above the true rate of $25.22\%$.
Consequently, H3 top-1 accuracy near the SIL prevalence does not establish useful procedural anticipation.
We therefore retain GTEA and EPIC-KITCHENS-100 as dataset-resolved calibration diagnostics rather than including them in the main aggregate comparison.
However, it is important to note that TRACE outperforms most of the other methods in activity classification while remaining close to black-box performance in forecasting.

\begin{table*}[h]
    \centering
    \caption{\textbf{GTEA and EPIC-KITCHENS-100 diagnostic comparison.} PE-L14, 128-concept results on GTEA s1 verb targets and EPIC-KITCHENS-100 s1 action targets, with activity classification, H3 forecast accuracy, and H3 SIL calibration reported over seeds 42--46.}
    \label{tab:gtea-epic-128-concept-diagnostics}
    \begingroup
    \scriptsize
    \setlength{\tabcolsep}{4pt}
    \resizebox{\textwidth}{!}{%
    \begin{tabular}{llcccccc}
          \toprule
          & & \multicolumn{2}{c}{Activity} & \multicolumn{2}{c}{H3} & \multicolumn{2}{c}{H3 SIL
          diagnostics} \\
          \cmidrule(lr){3-4} \cmidrule(lr){5-6} \cmidrule(lr){7-8}
          Dataset & Method & Top-1 & Top-3 & Top-1 & Top-3 & Predicted rate & False-positive rate \\
          \midrule
          \multirow{5}{*}{GTEA}
          & Sparse Linear
          & $\mval{21.7}{2.5}$ & $\mval{41.5}{3.3}$ & $\bmval{49.3}{0.4}$ & \underline{$\mval{62.5}
          {0.5}$} & $\mval{98.02}{0.85}$ & $\mval{97.30}{0.90}$ \\
          & MoTIF
          & $\bmval{49.5}{0.1}$ & \underline{$\mval{62.1}{0.4}$} & n/a & n/a & n/a & n/a \\
          & TRACE (ours)
          & \underline{$\mval{49.2}{0.2}$} & $\bmval{62.5}{0.2}$ & $\mval{45.1}{0.6}$ & $\mval{61.7}
          {0.4}$ & $\mval{85.51}{1.56}$ & $\bmval{82.62}{1.76}$ \\
          & Feature Transformer
          & $\mval{45.5}{2.3}$ & $\mval{61.3}{0.8}$ & $\mval{47.6}{1.9}$ & $\mval{61.9}{0.7}$ & $
          \mval{92.90}{3.61}$ & \underline{$\mval{91.25}{3.31}$} \\
          & SlowFast-style
          & $\mval{46.7}{1.6}$ & \underline{$\mval{62.1}{0.9}$} & \underline{$\mval{48.2}{0.6}$} & $
          \bmval{62.6}{0.4}$ & $\mval{94.03}{2.38}$ & $\mval{92.59}{3.07}$ \\
          \midrule
          \multirow{5}{*}{EPIC-KITCHENS-100}
          & Sparse Linear
          & $\mval{24.1}{3.0}$ & $\mval{33.4}{3.2}$ & $\mval{25.33}{0.18}$ & $\mval{36.4}{0.2}$ & $
          \mval{73.89}{0.89}$ & $\mval{71.24}{0.98}$ \\
          & MoTIF
          & $\mval{24.7}{0.4}$ & $\mval{36.5}{0.3}$ & n/a & n/a & n/a & n/a \\
          & TRACE (ours)
          & $\bmval{26.6}{0.2}$ & $\bmval{39.0}{0.2}$ & $\mval{25.09}{0.18}$ & $\bmval{37.0}{0.3}$ & $
          \mval{66.70}{0.35}$ & $\bmval{63.10}{0.34}$ \\
          & Feature Transformer
          & $\mval{26.3}{0.3}$ & $\mval{36.8}{0.5}$ & \underline{$\mval{25.34}{0.22}$} & $\mval{35.3}
          {0.4}$ & $\mval{67.08}{1.69}$ & \underline{$\mval{63.54}{1.76}$} \\
          & SlowFast-style
          & \underline{$\mval{26.5}{1.1}$} & \underline{$\mval{38.6}{0.9}$} & $\bmval{25.68}{0.60}$ &
          \underline{$\mval{36.7}{0.3}$} & $\mval{69.15}{3.79}$ & $\mval{65.61}{4.03}$ \\
          \bottomrule
    \end{tabular}%
    }
    \endgroup
\end{table*}

\subsection{Additional dataset protocols}
\label{app:all-protocols}

Table~\ref{tab:all-protocols} reports TRACE on every benchmark protocol used in the five-seed evaluation.
The main paper uses the BARISTA, averages Breakfast official s1--s4 within each seed, and uses the MPII Attr protocol to match the principal comparison.
For BARISTA, which provides only a train/test partition, we use a deterministic video-level split that orders the 185 videos by the SHA-1 hash of their identifiers and assigns the first 70\% to training, the next 15\% to validation, and the remaining 15\% to testing (129/28/28 videos).
The exact assignment is provided with the released code.
The per-protocol rows below, the MPII dish protocol, GTEA Gaze, and EPIC-KITCHENS-100 provide additional cross-protocol evidence.
GTEA uses verb targets~\citep{li2021eye}, whereas EPIC-KITCHENS-100 uses composite verb--noun action targets~\citep{damen2020epic}.

\begin{table}[h]
    \centering
    \caption{\textbf{Proposed-model protocol sweep.} PE-L14 proposed-model results with 128-concept vocabularies on the individual official protocols (BARISTA, Breakfast s1--s4, MPII Attr/Dishes, GTEA s1 verb, and EPIC-KITCHENS-100 s1 action), reported over seeds 42--46.
    Entries are mean $\pm$ standard deviation over five seeds in percent.
    The matched PE-L14 multi-method matrix in Table~\ref{tab:blackbox-concept-baselines} is a different run, so its entries are not expected to match these protocol-specific values.}
    \label{tab:all-protocols}
    \begingroup
    \scriptsize
    \setlength{\tabcolsep}{3pt}
    \begin{tabular}{llrrrr}
        \toprule
        Dataset & Protocol & \multicolumn{2}{c}{Activity} & \multicolumn{2}{c}{H3} \\
        \cmidrule(lr){3-4} \cmidrule(lr){5-6}
        & & Top-1 & Top-3 & Top-1 & Top-3 \\
        \midrule
        BARISTA & video-level & $\mval{76.6}{1.2}$ & $\mval{89.0}{0.5}$ & $\mval{67.3}{0.3}$ & $\mval{81.2}{0.5}$ \\
        Breakfast & official s1 & $\mval{50.4}{0.8}$ & $\mval{75.3}{0.6}$ & $\mval{43.8}{0.5}$ & $\mval{67.8}{0.7}$ \\
        Breakfast & official s2 & $\mval{35.9}{0.6}$ & $\mval{60.4}{0.4}$ & $\mval{29.3}{0.7}$ & $\mval{52.4}{0.6}$ \\
        Breakfast & official s3 & $\mval{44.0}{0.7}$ & $\mval{69.4}{0.4}$ & $\mval{37.4}{0.7}$ & $\mval{61.2}{1.0}$ \\
        Breakfast & official s4 & $\mval{47.8}{0.8}$ & $\mval{73.8}{0.8}$ & $\mval{41.9}{0.8}$ & $\mval{67.9}{0.5}$ \\
        MPII Cooking~2 & Attr & $\mval{46.5}{0.5}$ & $\mval{67.3}{0.4}$ & $\mval{39.0}{0.3}$ & $\mval{58.7}{0.3}$ \\
        MPII Cooking~2 & Dishes & $\mval{45.1}{0.4}$ & $\mval{67.5}{0.6}$ & $\mval{37.3}{0.6}$ & $\mval{58.1}{0.6}$ \\
        GTEA Gaze & verb & $\mval{49.4}{0.1}$ & $\mval{62.6}{0.3}$ & $\mval{48.3}{0.5}$ & $\mval{62.7}{0.1}$ \\
        EPIC-KITCHENS-100 & action & $\mval{26.7}{0.2}$ & $\mval{38.8}{0.3}$ & $\mval{25.9}{0.2}$ & $\mval{36.9}{0.3}$ \\
        \bottomrule
    \end{tabular}%
    \endgroup
\end{table}

Performance varies substantially across official folds and target surfaces.
For example, Breakfast H3 top-1 ranges from $29.3\%$ on official s2 to $43.8\%$ on official s1.
The GTEA and EPIC-KITCHENS results should also be interpreted together with their large unlabeled regions and SIL calibration, rather than compared directly with the main action-recognition results.

\subsection{Synthetic dataset evaluation}
\label{app:synthetic-delayed}

Table~\ref{tab:synthetic-delayed} reports a controlled sequence task containing six known delayed concept relations, the concrete implementation of the synthetic dataset is provided in the supplementary material. 
The experiment isolates whether the temporal graph can exploit a dependency that is unavailable to a current-only model and difficult to represent with flat sparse dynamics.

The temporal graph reaches $77.6\%$ H3 accuracy, compared with $6.6\%$ when the graph is disabled and $7.6\%$ for sparse linear dynamics.
It also retrieves $83.3\%$ of the planted relations among its active forecast edges.
Precision is lower at $19.8\%$, showing that predictive success does not imply exact recovery of the generating graph.
The experiment establishes that the implementation can exploit and partially identify delayed structure under controlled conditions, but it does not establish causal recovery on real videos.

\begin{table}[h]
    \centering
    \caption{\textbf{Synthetic delayed-dependency diagnostic.} Six-relation synthetic delayed-dependency task, trained with seeds 42--44.
    Activity accuracies, active-edge recall, and active-edge precision are reported in percent.}
    \label{tab:synthetic-delayed}
    \begingroup
    \scriptsize
    \setlength{\tabcolsep}{3pt}
    \begin{tabular}{lrrrr}
        \toprule
        Model & H1 & H3 & Edge recall & Edge precision \\
        \midrule
        Temporal graph & $\mval{75.0}{4.0}$ & $\mval{77.6}{1.0}$ & $\mval{83.3}{16.7}$ & $\mval{19.8}{4.3}$ \\
        History, graph disabled & $\mval{6.6}{0.2}$ & $\mval{6.6}{0.6}$ & $\mval{0.0}{0.0}$ & $\mval{0.0}{0.0}$ \\
        Current-only, graph disabled & $\mval{6.6}{0.2}$ & $\mval{6.6}{0.6}$ & $\mval{0.0}{0.0}$ & $\mval{0.0}{0.0}$ \\
        Sparse linear dynamics & $\mval{7.6}{0.2}$ & $\mval{7.6}{0.2}$ & n/a & n/a \\
        \bottomrule
    \end{tabular}%
    \endgroup
\end{table}

\section{Graph properties}
\label{app:graph-properties}

\subsection{Effective rollout dependency concentration}
\label{app:jacobian-dependency}

To assess how localized the learned rollout dependencies are, we compute the absolute rollout Jacobian from the input concept state to the predicted concept state, normalize parent contributions, and summarize the resulting influence distribution over eight held-out windows per dataset and seed.
Table~\ref{tab:jacobian-dependency} reports the effective number of influential parents ($N_{\mathrm{eff}}$), cumulative influence in the five strongest parents (Top-5), same-concept influence (Diag), effective density above a 1\% influence threshold, and the number of parents needed to explain 90\% of influence.

\begin{table*}[h]
    \centering
    \caption{\textbf{Effective rollout dependency concentration.} PE-L14, 128-concept graph and dense-forecast-rollout checkpoints on BARISTA, Breakfast s1, and MPII Attr, evaluated on eight held-out windows per dataset and seed for seeds 42--44. Each entry is graph $\rightarrow$ dense. $N_{\mathrm{eff}}$ and 90\% support are parent counts; Top-5, Diag, and Density are percentages. Density is the fraction of dependencies carrying at least 1\% of an output's normalized influence, not the structural connectivity of the dense model.}
    \label{tab:jacobian-dependency}
    \begingroup
    \scriptsize
    \setlength{\tabcolsep}{4pt}
    \begin{tabular}{llccccc}
        \toprule
        Dataset & Horizon & $N_{\mathrm{eff}}$ & Top-5 ($\%$) & Diag ($\%$) & Density ($\%$) & 90\% support \\
        \midrule
        \multirow{2}{*}{BARISTA}
        & H1 & $1.9 \rightarrow 72.7$ & $93.0 \rightarrow 23.9$ & $90.7 \rightarrow 15.6$ & $1.1 \rightarrow 23.5$ & $2.3 \rightarrow 79.8$ \\
        & H3 & $4.1 \rightarrow 89.1$ & $82.6 \rightarrow 16.2$ & $76.9 \rightarrow 6.7$ & $4.8 \rightarrow 27.8$ & $13.8 \rightarrow 81.8$ \\
        \midrule
        \multirow{2}{*}{Breakfast s1}
        & H1 & $2.4 \rightarrow 77.4$ & $89.7 \rightarrow 21.8$ & $85.9 \rightarrow 13.1$ & $2.3 \rightarrow 24.7$ & $5.7 \rightarrow 80.1$ \\
        & H3 & $6.2 \rightarrow 90.7$ & $75.0 \rightarrow 15.3$ & $66.5 \rightarrow 5.7$ & $10.8 \rightarrow 28.4$ & $16.2 \rightarrow 82.1$ \\
        \midrule
        \multirow{2}{*}{MPII Attr}
        & H1 & $2.2 \rightarrow 69.6$ & $90.7 \rightarrow 25.5$ & $87.3 \rightarrow 17.3$ & $1.9 \rightarrow 22.6$ & $4.9 \rightarrow 78.8$ \\
        & H3 & $5.5 \rightarrow 86.6$ & $77.3 \rightarrow 17.4$ & $69.5 \rightarrow 7.7$ & $9.7 \rightarrow 27.2$ & $14.9 \rightarrow 81.1$ \\
        \bottomrule
    \end{tabular}%
    \endgroup
\end{table*}

The dense rollout is substantially more entangled by every concentration measure.
At H3, its effective parent count is $86.6$--$90.7$, compared with $4.1$--$6.2$ for the graph rollout, only $15.3$--$17.4\%$ of dense influence lies in the five strongest parents, compared with $75.0$--$82.6\%$ for the graph.
Same-concept influence falls to $5.7$--$7.7\%$ in the dense model, so more than 92\% of its effective influence is off-diagonal, whereas the graph retains $66.5$--$76.9\%$ same-concept influence.
The dense rollout also requires $81.1$--$82.1$ parents to explain 90\% of influence, versus $13.8$--$16.2$ for the graph.
Thus, the dense arm is not merely harder to visualize: its learned dependency structure is less concentrated and less concept-local.
The reported density is effective Jacobian density at the 1\% threshold, the dense structural connectivity remains 100\%.

\subsection{Graph robustness and limits of causal interpretation}
\label{app:graph-stability}

Table~\ref{tab:graph-stability} compares the active-learned edge sets between all pairs of seeds for each main dataset.
Jaccard overlap measures whether the same edges are selected, while sign agreement is computed only on edges active in both seeds.

\begin{table}[h]
    \centering
    \caption{\textbf{Pairwise graph stability.} PE-L14, 128-concept graph stability on the BARISTA, Breakfast s1, and MPII Attr protocols, computed across seeds 42--44.
    Values are means across seed pairs in percent.}
    \label{tab:graph-stability}
    \begingroup
    \scriptsize
    \setlength{\tabcolsep}{6pt}
    \begin{tabular}{lrr}
        \toprule
        Dataset & Edge-set Jaccard & Shared-edge sign agreement \\
        \midrule
        BARISTA & $5.3$ & $88.5$ \\
        Breakfast & $11.9$ & $98.8$ \\
        MPII Cooking~2 Attr & $11.7$ & $98.9$ \\
        \bottomrule
    \end{tabular}
    \endgroup
\end{table}

The exact selected edge sets have low overlap across seeds, especially on BARISTA.
However, edges that recur across seeds usually retain their sign.
The learned graphs therefore provide seed-specific sparse explanations with relatively consistent directions among shared relations, rather than a unique recovered graph.
The synthetic experiment above tests known delayed dependencies, whereas the real-video stability analysis is a reproducibility diagnostic, neither identifies causal structure in the video-generating process.

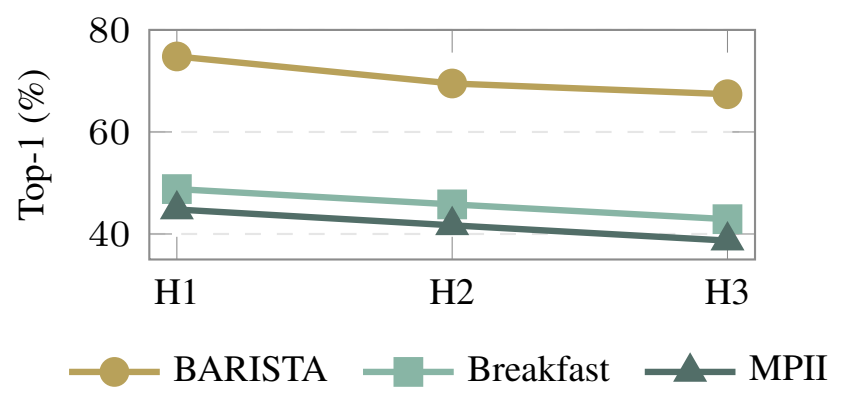
\begin{wrapfigure}{r}{0.42\textwidth}
    \centering
    \vspace{-1.4em}

    \begin{tikzpicture}
    \begin{axis}[
        width=0.4\textwidth,
        height=3.1cm,
        xmin=0.9, xmax=3.1,
        ymin=35, ymax=80,
        xtick={1,2,3},
        xticklabels={H1,H2,H3},
        ylabel={Top-1 (\%)},
        axis line style={draw=black!45},
        tick style={draw=black!45},
        ymajorgrids,
        grid style={black!10, dashed},
        tick label style={font=\scriptsize},
        label style={font=\scriptsize},
        legend style={
            draw=none,
            fill=none,
            font=\scriptsize,
            at={(0.5,-0.32)},
            anchor=north,
            legend columns=3,
            /tikz/every even column/.append style={column sep=4pt}
        },
    ]

    \addplot[
        vlmgold,
        very thick,
        mark=*,
        mark size=2.2pt
    ] coordinates {
        (1,74.8) (2,69.5) (3,67.4)
    };
    \addlegendentry{BARISTA}

    \addplot[
        vlmgreen,
        very thick,
        mark=square*,
        mark size=2.2pt
    ] coordinates {
        (1,48.8) (2,45.8) (3,42.9)
    };
    \addlegendentry{Breakfast}

    \addplot[
        vlmdark,
        very thick,
        mark=triangle*,
        mark size=2.5pt
    ] coordinates {
        (1,44.8) (2,41.7) (3,38.7)
    };
    \addlegendentry{MPII}

    \end{axis}
    \end{tikzpicture}
    
    \caption{\textbf{Forecasting diagnostics.}
    Top-1 accuracy across forecast horizons H1-H3.}
    \label{fig:forecast-diagnostics}
    \vspace{-2em}
\end{wrapfigure}

\subsection{Forecasting diagnostics}
We evaluate H1--H3 forecasts to assess error accumulation under recursive rollout and the reliability of decoded activity confidences.
Top-1 accuracy declines consistently but moderately with increasing horizon (Fig. \ref{fig:forecast-diagnostics}).
BARISTA remains robust at H3 and exhibits comparatively well-aligned confidence estimates, whereas Breakfast and MPII are both less accurate and increasingly overconfident, motivating calibration-aware use of predicted probabilities in less predictable settings.

\subsection{Long-horizon rollout}
\label{app:long-horizon}

We test the same learned H3 transition recursively for up to 50 future windows without retraining it for the longer horizons.
This experiment therefore evaluates zero-shot rollout stability rather than direct H50 supervision.
Table~\ref{tab:long-horizon} reports activity top-1 accuracy at representative horizons and compares H50 with a persistence baseline that repeats the last observed activity with 500 test windows.

\begin{table*}[h]
    \centering
    \caption{\textbf{Zero-shot recursive rollout from H3 to H50.} PE-L14, 128-concept rollout on the BARISTA, Breakfast s1, and MPII Attr protocols, with entries reported over seeds 42--44 in percent.
    Persist. is the H50 last-observation baseline, and Sat. is the fraction of H50 concept activations near their range boundaries.}
    \label{tab:long-horizon}
    \begingroup
    \scriptsize
    \setlength{\tabcolsep}{4pt}
    \resizebox{\textwidth}{!}{%
    \begin{tabular}{lrrrrrrrr}
        \toprule
        Dataset & H3 & H10 & H20 & H30 & H40 & H50 & Persist. & Sat. \\
        \midrule
        BARISTA & $\mval{68.4}{0.2}$ & $\mval{60.0}{3.5}$ & $\mval{37.6}{22.8}$ & $\mval{31.2}{27.7}$ & $\mval{27.3}{29.6}$ & $\mval{25.5}{27.4}$ & $\mval{39.6}{0.0}$ & $\mval{56.9}{5.4}$ \\
        Breakfast & $\mval{45.7}{4.5}$ & $\mval{40.0}{1.9}$ & $\mval{36.2}{4.4}$ & $\mval{34.0}{4.1}$ & $\mval{31.3}{3.3}$ & $\mval{27.3}{2.7}$ & $\mval{32.7}{3.6}$ & $\mval{56.7}{1.6}$ \\
        MPII Cooking~2 Attr & $\mval{38.3}{1.9}$ & $\mval{21.1}{1.1}$ & $\mval{10.6}{1.1}$ & $\mval{10.7}{0.2}$ & $\mval{9.1}{2.6}$ & $\mval{9.5}{0.9}$ & $\mval{20.2}{1.7}$ & $\mval{52.9}{2.1}$ \\
        \bottomrule
    \end{tabular}
    }
    \endgroup
\end{table*}

Accuracy degrades with recursive depth on every dataset.
BARISTA remains relatively strong through H10, but its variance increases sharply beyond H20, indicating seed-dependent rollout instability.
At H50, TRACE falls below persistence on all three datasets and between $52.9\%$ and $56.9\%$ of concept activations are saturated.
The corresponding H50 concept mean absolute errors are $0.426$, $0.409$, and $0.414$ for BARISTA, Breakfast, and MPII Cooking~2 Attr.
Thus, the learned dynamics remain informative for moderately longer forecasts but do not support a claim of reliable 40--50-step prediction.

\section{Algorithmic description}
\label{app:algorithms}

\subsection{VLM concept proposal with the generation prompt}
\label{app:vlm-concepts}

Concept vocabulary construction is offline and label-blind at the visual-observation stage, following the broader direction of label-free and language-guided concept construction~\citep{oikarinen2023labelfree,yang2023language}.
Qwen3.5-27B \cite{qwenteam2026qwen35omnitechnicalreport} receives four-frame storyboards sampled in a label-balanced manner, returns structured visible objects, states, and interactions, and then proposes short reusable concepts.
Candidate phrases are deduplicated, rejected when they exactly match a normalized activity label, retrieved with text--image similarity, and retained only after image-grounded verification on at least two videos.
Each candidate includes supporting storyboard identifiers and a short factual visual rationale before retrieval and verification.
Concept embeddings are matched to training-window embeddings by cosine similarity in the shared vision--language space~\citep{clip,bolya2025perception}, which supplies the visual evidence used by the verification stage.
The storyboard analysis is label-blind, activity labels are attached only afterward for coverage auditing and for the proposal-stage instruction not to restate labels.
The compact prompt template below reproduces the operative constraints.

\begin{quote}\footnotesize\ttfamily
You are building a reusable visual concept vocabulary for dataset \emph{[dataset]}. Propose exactly \emph{[N]} short concepts grounded in the supplied label-blind storyboard observations. Each phrase must contain 1--3 words and be reusable across videos. Prefer concrete objects/materials, visible states/attributes, and spatial or hand-object relations. Include an action primitive only when motion is directly supported across the four frames. Do not repeat, restate, or semantically paraphrase an activity label. Do not use SIL/background as a concept. For every candidate, cite supporting storyboard IDs and give one factual rationale of at most 16 words. Return the required JSON schema only.
\end{quote}

\subsection{Training and rollout algorithms}
\label{app:algorithms-all}

The algorithms below summarize the two implementation-critical pipelines:
offline vocabulary construction and autoregressive concept rollout.

Algorithm~\ref{alg:vlm-concept-generation} gives the offline procedure for proposing, grounding, deduplicating, and exporting the textual concept vocabulary.

\begin{algorithm}[t]
\caption{\textbf{VLM-based textual concept generation.}}
\label{alg:vlm-concept-generation}
\begin{algorithmic}[1]
\STATE \textbf{Input:} training-window index with frozen PE-L14 features, Qwen3.5-27B, seed $s$, target size $C$
\STATE \textbf{Input:} proposal count $N$, round size $K$, number of rounds $R$, retrieval count $K_r$, verification threshold $V$, similarity threshold $\delta$
\STATE Initialize candidate set $\mathcal{Q}\leftarrow\varnothing$ and label counts
\FOR{$r=1,\ldots,R$}
    \STATE Sample $K$ balanced training windows with $\operatorname{Random}(s+r)$; render four-frame storyboards
    \STATE Parse storyboards with the label-blind VLM prompt; attach labels only for coverage auditing
    \STATE Propose $N$ grounded 1--3 word concepts with categories, evidence IDs, and rationales
    \STATE Reject malformed, overlong, duplicate, exact-label, unsupported, or invalid-evidence proposals
    \FOR{each accepted proposal $q$}
        \STATE Retrieve up to $K_r$ nearest windows from distinct videos using PE-L14 cosine similarity
        \STATE VLM-verify the retrieved windows; retain $q$ iff at least $V$ windows support it
    \ENDFOR\STATE Select grounded candidates by utility (verification confidence, visual support, category coverage, and brevity), remove text-embedding duplicates above $\delta$, and save round state
    \IF{$|\mathcal{Q}|\geq C$}
        \STATE \textbf{break}
    \ENDIF
\ENDFOR
\STATE Validate count, phrase length, uniqueness, label exclusion, grounding, and redundancy
\STATE \textbf{Output:} concept vocabulary $\mathcal{C}$ and text embeddings
\end{algorithmic}
\end{algorithm}
For the reported generation runs, $(C,N,K,R,K_r,V,\delta,s)=(128,80,24,12,6,2,0.95,42)$; incomplete proposal JSON is retried with a larger token budget.

Algorithm~\ref{alg:autoregressive-rollout} provides our method's procedure for
autoregressive forecasting, teacher forcing, and evaluation-time interventions.

\begin{algorithm}[t]
\caption{\textbf{Autoregressive concept rollout and training-time teacher forcing.}}
\label{alg:autoregressive-rollout}
\begin{algorithmic}[1]
\STATE \textbf{Input:} concept scores $\mathbf{c}_{1:T}$, validity mask
$\mathbf{m}_{1:T}$, graph modules $\mathcal{G}_{\rm shared}$,
$\mathcal{G}_{\rm act}$, $\mathcal{G}_{\rm fut}$, activity head
$(\mathbf{W},\mathbf{b})$, activity-feedback matrix $\mathbf{A}_{\rm act}$,
history length $L_{\rm hist}$, lag gates
$\{\lambda_\delta\}_{\delta=1}^{L_{\rm hist}}$, and horizon $H$
\STATE \textbf{Optional:} interventions $\mathcal{I}$, training labels
$y_{1:T+H}$, and teacher-forcing ratio $\rho$
\STATE $\mathbf{z}_{1:T}\leftarrow
\operatorname{Normalize}(\mathbf{c}_{1:T})\odot\mathbf{m}_{1:T}$;
apply interventions at their configured concept, edge, or activity-belief hook
\STATE $\mathbf{r}^{\rm shared}_{1:T}\leftarrow
\mathcal{G}_{\rm shared}(\mathbf{z}_{1:T})$
\STATE $\mathbf{r}^{\rm act}_{1:T}\leftarrow
\mathcal{G}_{\rm act}(\mathbf{r}^{\rm shared}_{1:T})$;
$\mathbf{r}^{\rm fut}_{1:T}\leftarrow\mathbf{r}^{\rm shared}_{1:T}$
\STATE Compute observed activity beliefs
$\hat{\mathbf{y}}_s\leftarrow
\operatorname{softmax}\!\left(
\phi(\mathbf{r}^{\rm act}_s)\mathbf{W}^{\top}+\mathbf{b}\right)$
for valid $s\leq T$
\STATE Compute $d_T^{\rm hist}$ using Eq.~(5), and obtain
$\widetilde{\mathbf{r}}_T^{\rm act}$ and
$\widetilde{\mathbf{r}}_T^{\rm fut}$ using Eq.~(6)
\STATE Set $\widetilde{\mathbf{r}}_s^\alpha\leftarrow\mathbf{r}_s^\alpha$
for $s<T$ and $\alpha\in\{\rm act,fut\}$
\STATE $\hat{\mathbf{y}}_T\leftarrow
\operatorname{softmax}\!\left(
\phi(\widetilde{\mathbf{r}}_T^{\rm act})\mathbf{W}^{\top}+\mathbf{b}
\right)$
\FOR{$h=1,\ldots,H$}
    \STATE $t\leftarrow T+h-1$
    \STATE During training, replace each available source belief
    $\hat{\mathbf{y}}_s$ with
    $\operatorname{TeacherForce}(
    \hat{\mathbf{y}}_s,\operatorname{onehot}(y_s);\rho)$;
    otherwise leave $\hat{\mathbf{y}}_s$ unchanged
    \STATE $d_{t+1}^{\rm hist}\leftarrow
    \sum_{\delta=1}^{L_{\rm hist}}
    \sigma(\lambda_\delta)m_{t+1-\delta}
    \hat{\mathbf{y}}_{t+1-\delta}\mathbf{A}_{\rm act}$
    \STATE $\widetilde{\mathbf{r}}_{t+1}^{\rm fut}\leftarrow
    \sigma\!\left(
    \phi\!\left(
    [\mathcal{G}_{\rm fut}(
    \widetilde{\mathbf{r}}_{1:t}^{\rm fut})]_{t+1}
    \right)+d_{t+1}^{\rm hist}\right)$
    \STATE $\hat{\mathbf{y}}_{t+1}\leftarrow
    \operatorname{softmax}\!\left(
    \phi(\widetilde{\mathbf{r}}_{t+1}^{\rm fut})\mathbf{W}^{\top}
    +\mathbf{b}\right)$;
    $m_{t+1}\leftarrow1$
\ENDFOR
\IF{training}
    \STATE Form stop-gradient future concept targets using Eq.~(1) and
    optimize $L$ from Eq.~(9), with detailed terms in Eqs.~(10)--(14)
\ENDIF
\STATE \textbf{Output:} current and future activity predictions, forecast
concept states, and intervention diagnostics
\end{algorithmic}
\end{algorithm}
The forecast graph module $\mathcal{G}_{\rm fut}$ is reused as a one-step transition for each autoregressive forecast step; persistence ablations replace it with the identity. 
Interventions are applied at the corresponding concept, activity-belief, or edge hook, and oracle edit ranking is used only for
evaluation.

\newpage
\section{Hyperparameters and reproducibility}
\label{app:implementation}

This section specifies the complete training configuration used for the main 128-concept comparison.
Unless stated otherwise, the same settings are used for Sparse Linear, MoTIF, the proposed model, the Feature Transformer, and the SlowFast-style baseline.
The 256-concept experiment changes only the dataset-specific concept vocabulary and the resulting concept dimensionality. The black-box models operate on the frozen visual features and are therefore independent of the concept budget.

\subsection{Dataset and baseline details}
\label{app:dataset-baseline-details}

\textbf{Datasets and targets.}
Breakfast and MPII Cooking~2 cover cooking and daily manipulation activities, whereas BARISTA focuses on fine-grained beverage preparation.
The five-seed main comparison uses the BARISTA, averages Breakfast s1--s4 and MPII Cooking~2 Attr within each seed, and evaluates action-label targets with SIL assigned to otherwise unlabeled windows; the dataset protocols follow the corresponding benchmark descriptions~\citep{knab2026barista,breakfast,rohrbach15ijcv}.
For BARISTA, which provides only a train/test partition, we use a deterministic video-level split that orders the 185 videos by the SHA-1 hash of their identifiers and assigns the first 70\% to training, the next 15\% to validation, and the remaining 15\% to testing (129/28/28 videos).
The exact assignment is provided with the released code.
GTEA Gaze uses verb targets~\citep{li2021eye} and EPIC-KITCHENS-100 uses composite verb--noun action targets~\citep{damen2020epic}, and their many unlabeled windows make them diagnostic stress tests rather than part of the main aggregate comparison.
Their true H3 SIL rates are $49.41\%$ and $25.22\%$, respectively, and their dataset-resolved results appear in Appendix~\ref{app:gtea_epic} and Appendix~\ref{app:all-protocols}.

\textbf{Perception Encoder video embeddings.}
All datasets are extracted once with pretrained Perception Encoder~\citep{bolya2025perception} backbones and are kept frozen.
We partition each video into non-overlapping 32-frame windows, and the final non-empty partial window is retained.
For each window, eight frames are sampled uniformly, including the temporal endpoints when the window contains at least eight frames, while shorter windows are completed by repeating their last frame.
The frames are converted to RGB and processed with the official PE image transform at the model's $336\times336$ input resolution, stacked as a $(8,3,336,336)$ clip, and passed to the model's video encoder.
The resulting 1024-dimensional clip feature is computed in evaluation mode without gradients.
The named concepts (in text) are embedded with the same PE text encoder and tokenizer.

\textbf{Baselines.}
The zero-shot baseline is training-free and predicts the activity label whose text prompt ``a video of {label}'' has the highest cosine similarity to each frozen video-window feature, using the same vision--language embedding principle as CLIP~\citep{clip} and PE-L14~\citep{bolya2025perception}.
It uses no concept vocabulary, learned activity head, or temporal forecasting module, and therefore reports current-activity top-1/top-3 accuracy only.
Sparse Linear uses the proposed model's shared activity predictor with a sparse linear concept-dynamics head whose sparsity budget matches the active graph-edge count.
MoTIF is a concept-based activity-recognition baseline without a comparable concept-rollout H3 endpoint under our protocol~\citep{knab2026concepts}.
The SlowFast-style temporal convolutional model and feature transformer operate directly on the frozen video-window features before concept projection, without concept activations or graph structure~\citep{feichtenhofer2019slowfast,bertasius2021space}.
Thus, the black-box baselines test opaque temporal modeling over the same visual evidence, while Sparse Linear tests whether sparse concept dynamics alone explain the graph-based model's forecasting behavior.

\subsection{Further implementations details}

\subsubsection{Shared optimization settings}

Table~\ref{tab:repro-shared-hparams} lists the shared optimization and loss hyperparameters.
Models are trained with Adam without a learning-rate scheduler.
The checkpoint with the highest validation H3 top-3 accuracy is retained, with early stopping after 30 epochs without improvement.
Horizon losses are averaged uniformly over H1--H3.
Activity and forecast supervision use cross-entropy, while future concept supervision uses Smooth L1 loss.
The SIL false-positive terms add a penalty whenever SIL is predicted for a non-SIL target.

\begin{table*}[h]
    \centering
    \caption{\textbf{Shared optimization and loss settings.} Shared optimization, sampling, and loss hyperparameters that are active in the reported runs.}
    \label{tab:repro-shared-hparams}
    \begingroup
    \scriptsize
    \setlength{\tabcolsep}{5pt}
    \resizebox{\textwidth}{!}{%
    \begin{tabular}{ll@{\hspace{18pt}}ll}
        \toprule
        Hyperparameter & Value & Hyperparameter & Value \\
        \midrule
        Optimizer & Adam & Learning rate & $1\times10^{-4}$ \\
        Weight decay & $1\times10^{-4}$ & Batch size & 256 \\
        Maximum epochs & 200 & Early-stopping patience & 30 epochs \\
        Selection metric & Validation H3 top-3 & Random seeds & 42, 43, 44, 45, 46 \\
        TF32 & Enabled & Observed history & 5 windows \\
        Forecast horizons & H1, H2, H3 & Frames per window & 32 \\
        Window sampling & Random training windows & Training sampler & Uniform \\
        Activity loss & Cross-entropy & Forecast loss & Cross-entropy \\
        Activity SIL-FP penalty & 0.5 & Forecast SIL-FP penalty & 0.5 \\
        Classifier $\ell_1$ weight & $10^{-3}$ & Horizon loss weights & Uniform \\
        \bottomrule
    \end{tabular}%
    }
    \endgroup
\end{table*}

\subsubsection{Method-specific architectures}

Table~\ref{tab:repro-model-hparams} reports all method-specific parameters outside the shared configuration.
Train-split standardization is fitted separately for the Sparse Linear and raw-feature black-box inputs, and the resulting statistics are applied unchanged to validation and test data.

\begin{table*}[h]
    \centering
    \caption{\textbf{Method-specific architecture settings.} Method-specific architecture and training hyperparameters for the main comparison.}
    \label{tab:repro-model-hparams}
    \begingroup
    \scriptsize
    \setlength{\tabcolsep}{5pt}
    \begin{tabular}{lp{0.76\textwidth}}
        \toprule
        Method & Hyperparameters \\
        \midrule
        Sparse Linear
        & Shared activity/forecast predictor; sparse concept-dynamics matrix; top-$k=20$ incoming dynamics connections per target concept; future-concept Smooth L1 weight 0.5. \\
        MoTIF
        & One per-concept temporal transformer layer; channel width/dimension 1; dropout 0.1; shared activity head enabled; transition-tolerant forecast radius 1 with mixing weight 0.25. The paper reports its activity endpoint only because its forecast is not a comparable explicit concept rollout. \\
        Feature Transformer
        & Input projection to 256 dimensions; two Transformer encoder layers; four attention heads; feed-forward dimension 512; GELU activation; dropout 0.1; learned positional embedding over the five observed windows; last valid token pooling. \\
        SlowFast-style
        & Input projection to 256 dimensions; two fast temporal-convolution layers; kernel size 3; GELU activation; dropout 0.1; slow-path stride 2; last-valid fast pooling and masked-mean slow pooling; concatenated 512-dimensional representation. \\
        Proposed model
        & Graph and rollout settings are enumerated in Table~\ref{tab:repro-proposed-hparams}; all parameters are selected during the single joint-training run. \\
        \bottomrule
    \end{tabular}
    \endgroup
\end{table*}

Both black-box architectures use one activity head and separate linear forecast heads for H1--H3.
Their inputs are the same frozen window features used upstream of concept projection in the concept-based models.

\subsubsection{Graph and rollout settings}

The complete active and implementation-default configuration for TRACE is given in Table~\ref{tab:repro-proposed-hparams}.
The observed concept sequence is refined by the graph, and future concepts are then recursively forecast for H1--H3.

\begin{table*}[h]
    \centering
    \caption{\textbf{Proposed-model graph and rollout settings.} Graph, rollout, and feedback hyperparameters for the proposed model.}
    \label{tab:repro-proposed-hparams}
    \begingroup
    \scriptsize
    \setlength{\tabcolsep}{5pt}
    \resizebox{\textwidth}{!}{%
    \begin{tabular}{ll@{\hspace{18pt}}ll}
        \toprule
        Hyperparameter & Value & Hyperparameter & Value \\
        \midrule
        Future concept horizons & 1, 2, 3 & Representation & Refined concept state only \\
        Concept forecast loss & Smooth L1, weight 0.5 & Concept-loss dead zone & 0 standard deviations \\
        Teacher forcing schedule & $0.5\rightarrow0$ linearly & Transition-tolerant loss & Radius 1, weight 0.25 \\
        Observed refiner & Graph & Forecast rollout mode & Fixed-length recursive \\
        Shared graph layers & 1 & Task-specific graph layers & 1 \\
        Spatial top-$k$ & 20 & Spatial soft threshold & 0 \\
        Same-concept temporal edges & Enabled & Temporal top-$k$ & 20 \\
        Temporal soft threshold & 0 & Cross-temporal edges & Enabled \\
        Cross-temporal top-$k$ & 20 & Cross-temporal threshold & 0 \\
        Edge-gate initialization & 0 & Edge reporting threshold & 0.2 \\
        Residual-gate initialization & 0 & Message scale & 1 \\
        State activation & Bounded logit & Prediction transform & Logit \\
        Forecast context & Flat ordered history & Activity context & Flat ordered history \\
        Video-level pooling & None & Past-context mode & None \\
        Motif-$z$ attention layers & 0 & Concept-score normalizer & Enabled \\
        Top-$k$ training mode & Gradual & Warm-up / ramp & 20 / 30 epochs \\
        Activity feedback mode & Sparse label-to-concept & Feedback top-$k$ & 40 \\
        Feedback gate initialization & 0 & Feedback history & 3 steps \\
        History-gate initialization & 0 & Feedback hidden dimension & 64 \\
        Feedback ridge coefficient & 0.1 & Probability epsilon & $10^{-4}$ \\
        Prototype smoothing & 1 & Transition prior & Disabled \\
        \bottomrule
    \end{tabular}%
    }
    \endgroup
\end{table*}

\subsection{Runtime and complexity}
\label{app:runtime-complexity}

Frozen window embeddings are computed once before model training and can be reused across models.
For a sequence of $N$ observed windows, an $H$-step forecast, $C$ concepts, $K$ activity classes, and $l$ graph message-passing layers applied per rollout step, the initial shared refinement and current-activity decoding cost $O(lNC^2+NKC)$ time.
Historical activity feedback adds at most $O(L_{\rm hist}KC)$ time.

The current implementation stores graph and activity-feedback matrices as dense tensors and applies top-$k$ masks to their entries.
Thus, the masks control the active learned messages but do not reduce the asymptotic cost of the dense matrix multiplications.
At every forecast step, the rollout applies $l$ graph layers and the shared activity head to the full $N$-slot state.
The $H$-step rollout therefore costs
\[
O\!\left(HN\left(lC^2+KC\right)\right),
\]
where the $KC$ term includes both activity decoding and activity-to-concept feedback across all $N$ positions.
We use $l=1$ and $L_{\rm hist}=3$ in the reported configuration.

The implementation retains the full concept-state and activity-belief sequence at every rollout step, requiring $O(HN(C+K))$ additional storage.
Vocabulary generation is an offline, one-time VLM and retrieval procedure and is not part of forecasting-time cost.
All costs are reported per video, for a batch of $B$ videos, multiply time costs by $B$.
We report asymptotic costs and measured latencies, and all experiments were conducted on NVIDIA RTX PRO 6000 GPUs.
Measured on an H200 (100 windows, five seeds), PE-L14 extraction takes 19\,ms per window and a TRACE H1 / H3 / H10 forecast 2.7 / 3.7 / 7.5\,ms (Sparse Linear, Feat.\ Transf., SlowFast: 0.10 / 0.34 / 0.23\,ms at H3), using 34--40\,MB of GPU memory. A concept intervention with five edits takes about 25\,ms. The latency is dominated by kernel-launch overhead, not computation.

\subsection{Concept-vocabulary reproducibility audit}
\label{app:concept-audit}

Table~\ref{tab:concept-audit} audits each main dataset's fixed vocabulary before interpreting downstream graph or intervention results.
The audit checks normalized textual duplicates, empirical activation prevalence, direct lexical overlap with target labels, and automated semantic-target warnings from the generation pipeline.

\begin{table}[h]
    \centering
    \caption{\textbf{128-concept vocabulary audit.} Automated audit of the 128-concept vocabularies.
    Lexical overlap and semantic warnings are fractions of concepts in percent.
    Degenerate concepts are near-zero or near-always-active concepts.}
    \label{tab:concept-audit}
    \begingroup
    \scriptsize
    \setlength{\tabcolsep}{4pt}
    \begin{tabular}{lrrrr}
        \toprule
        Dataset & Duplicates & Degenerate & Lexical overlap & Semantic warnings \\
        \midrule
        BARISTA & $0$ & $0$ & $0.8$ & $7.8$ \\
        Breakfast & $0$ & $0$ & $0.0$ & $1.6$ \\
        MPII Cooking~2 Attr & $0$ & $0$ & $0.0$ & $5.5$ \\
        \bottomrule
    \end{tabular}%
    \endgroup
\end{table}

The audit finds no normalized duplicates and no concepts with degenerate empirical prevalence.
Direct lexical target overlap is absent on Breakfast and MPII Cooking~2 Attr and occurs for one BARISTA concept.
The automated semantic check flags a small minority of concepts for possible target leakage or ambiguity, with the highest rate on BARISTA.
These checks support vocabulary hygiene, but frozen-encoder activations and VLM verification are automated evidence rather than human concept ground truth.

\section{Detailed training objective}
\label{app:training-objective}

\subsection{Full loss terms}

This appendix provides the full definitions of the loss terms summarized in
Section~\ref{sec:optimization}. Let
$\hat{\mathbf y}_{T+h}=\operatorname{softmax}(\boldsymbol{\ell}_{T+h})$,
with $\hat y_{T+h,a}$ denoting the predicted probability of activity $a$.
\begin{align}
\mathcal{L}_{\mathrm{act}}
&=
\operatorname{CE}(\boldsymbol{\ell}_T,y_T)
+\beta_{\mathrm{SIL}}\mathcal{L}_{\mathrm{SIL},T}, \\
\mathcal{L}_{\mathrm{forecast}}
&=
\sum_{h=1}^{H}\alpha_h
\Big[
(1-\gamma_{\mathrm{tol}})
\operatorname{CE}(\boldsymbol{\ell}_{T+h},y_{T+h})
+\gamma_{\mathrm{tol}}\mathcal{L}^{\mathrm{tol}}_h
+\beta_{\mathrm{SIL}}\mathcal{L}_{\mathrm{SIL},T+h}
\Big], \\
\mathcal{L}_{\mathrm{concept}}
&=
\sum_{h=1}^{H}\alpha_h
\operatorname{SmoothL1}\!\left(
\widetilde{\mathbf{r}}^{\,\mathrm{fut}}_{T+h},
\mathbf{z}^{\mathrm{target}}_{T+h}
\right), \\
\mathcal{L}^{\mathrm{tol}}_h
&=
-\log\sum_{a\in\mathcal{A}^{(r)}_h}\hat y_{T+h,a}, \\
\mathcal{A}^{(r)}_h
&=
\{y_{T+h},\ldots,y_{T+\min(H,h+r)}\}.
\end{align}

\subsection{Intervention protocol and policy-reproduction details}
\label{app:intervention-details}

The intervention evaluation covers BARISTA, Breakfast s1, and MPII Cooking~2 Attr, using seeds 42--44, 30 initially correct cases, and 30 initially incorrect H3 cases for every dataset--seed pair. The evaluation follows established concept-intervention protocols while extending them to recursive forecasts~\citep{koh2020concept,shin2023intervention,zarlenga2023intcem}.
The full evaluation records true-class probability change, label-flip rate, wrong-to-correct rate, correct-to-wrong rate, and future-concept displacement for each cumulative prefix.
Table~\ref{tab:intervention-interfaces} reports the primary interface comparison.

\subsubsection{Edit semantics and oracle information}

All edits are applied only at evaluation time to the frozen checkpoint, and each reported prefix is re-rolled out through H3 before scoring.
For a concept-node edit, the oracle ranks candidate concept coordinates by their directional contribution to the held-out H3 class relative to the predicted H1 state.
It then clamps each selected coordinate of the H1 rollout state to the corresponding normalized concept target extracted from the held-out next video window, which lies in $[0,1]$.
For an activity-belief edit, the oracle uses the ground-truth activity at one of the three preceding source steps $t$, $t+1$, or $t+2$ and sets its probability to $0.9$.
The remaining probability mass is redistributed proportionally among the other classes before the sparse activity-to-concept message is applied.
For an edge edit, the evaluator selects active learned relations by effective strength and multiplies each selected relation by $-1$ during the forward pass.
Consequently, edge inversion reverses the additive relation message but leaves the residual state path intact.
These definitions make all reported oracle results upper bounds on interface sensitivity rather than evidence that a user can identify the required edit from observed video alone.

Activity-belief interventions are selected from the three source steps that can affect H3.
At the maximum feasible activity budget, oracle-ranked edits increase true-class probability by $+3.0\pp$, $+28.3\pp$, and $+21.1\pp$ on BARISTA, Breakfast, and MPII, respectively.
Oracle-ranked concept-node edits at budget five increase the same quantity by $+7.2\pp$, $+11.3\pp$, and $+9.8\pp$.
Strongest learned-edge inversion changes true-class probability by no more than $0.02\pp$.
This small direct effect is consistent with the local scope of the edit: one relation message is reversed while the residual path and other active relations remain intact. Edge inversion therefore tests local relation sensitivity, while the intervention results overall establish sensitivity of the learned prediction mechanism rather than a causal effect in the video-generating process.

\subsubsection{Final rates, information access, and matched baselines}
\label{app:final-rates}

Table~\ref{tab:final-rates} reports final W$\rightarrow$C after the last edit (five seeds) by the information the policy uses. For the concept oracle, final and cumulative rates differ by at most $0.7\pp$. Confirming the current activity alone already corrects 35\% / 23\% of wrong forecasts on Breakfast / MPII (the automatic policies do not yet select helpful edits).

\begin{table}[h]
    \centering
    \caption{\textbf{Final correction rates by available information.} H3 W$\rightarrow$C (\%) after the last edit, five seeds.}
    \label{tab:final-rates}
    \scriptsize
    \begin{tabular}{lccc}
        \toprule
        Policy (information) & BARISTA & Breakfast s1 & MPII Attr \\
        \midrule
        Observed concepts only, 5 edits & 2.7 & 1.3 & 4.0 \\
        Blind Qwen (Table~\ref{tab:qwen-intervention-planning}) & 10.0 & 13.3 & 12.2 \\
        Current activity only, 1 belief edit & 3.3 & 35.3 & 23.3 \\
        Current and next two activities, 3 belief edits & 8.0 & 63.3 & 46.0 \\
        Concept oracle (future targets), 5 edits & 20.0 & 30.7 & 28.0 \\
        \bottomrule
    \end{tabular}
\end{table}

Table~\ref{tab:matched-interventions} applies the same oracle node edits (budget 5) to simpler editable models on the same cases. TRACE corrects 2--16$\times$ more wrong forecasts, harms none, and moves the non-edited concepts 2--16$\times$ less. For belief edits, the exact-capacity dense model is on par (65.6 vs.\ 63.3\% on Breakfast).

\begin{table}[h]
    \centering
    \caption{\textbf{Matched node edits.} Final W$\rightarrow$C / C$\rightarrow$W (\%) and mean shift of non-edited concepts; BARISTA / Breakfast s1 / MPII.}
    \label{tab:matched-interventions}
    \scriptsize
    \begin{tabular}{lccc}
        \toprule
        Model & W$\rightarrow$C & C$\rightarrow$W & Shift \\
        \midrule
        TRACE & 20.0 / 30.7 / 28.0 & 0 / 0 / 0 & .003 / .010 / .007 \\
        Sparse Linear & 10.1 / 13.7 / 3.2 & 0.6 / 1.4 / 0.0 & .042 / .039 / .026 \\
        Dense rollout & 8.1 / 6.5 / 5.2 & 2.2 / 4.5 / 2.8 & .022 / .020 / .023 \\
        Dense, exact capacity & 6.9 / 4.3 / 1.7 & 2.2 / 3.1 / 2.8 & .022 / .017 / .020 \\
        \bottomrule
    \end{tabular}
\end{table}

\subsubsection{Steering split by interface}
\label{app:steering-split}

Table~\ref{tab:steering-split} repeats Table~\ref{tab:non-oracle-steering} per interface on the same 900 cases (five seeds). Belief edits explain most of the success, concept edits alone reach up to 64\% when ranked by their measured effect on the target in the model, but stay at random level (0.7 / 7.3 / 7.7\%) when chosen by the heuristic or Qwen. Note that Qwen and the vocabulary generator share a model family. Example (Breakfast, target \textit{fry\_egg}): the heuristic sets the belief at $t$, $t{+}1$, $t{+}2$ and then switches \textit{oil bottle} on at $t{+}1$, $t{+}2$.

\begin{table}[h]
    \centering
    \caption{\textbf{Steering split by interface.} Target top-1 (\%); BARISTA / Breakfast s1 / MPII.}
    \label{tab:steering-split}
    \scriptsize
    \begin{tabular}{lccc}
        \toprule
        Policy & Concept only (5) & Belief only (3) & Both (5) \\
        \midrule
        Target heuristic & 0.7 / 10.3 / 4.0 & 4.7 / 90.0 / 85.3 & 4.0 / 88.0 / 83.7 \\
        Qwen & 0.7 / 12.3 / 5.3 & 4.7 / 90.0 / 85.3 & 7.7 / 82.7 / 78.0 \\
        Model-guided search & 10.7 / 64.0 / 58.3 & 8.0 / 90.3 / 86.3 & 20.0 / 93.7 / 94.0 \\
        \bottomrule
    \end{tabular}
\end{table}

\subsubsection{Qwen intervention prompt and visible context}
\label{app:qwen-prompt}

Each request contains one user message with an attached four-frame JPEG storyboard ending at time $t$ and the following template. 
Bracketed fields are instantiated for each case and the candidate-edit list is provided in full.

\begin{quote}
\footnotesize\ttfamily
You are an intervention policy for a concept-based video forecasting model.
The storyboard contains only the final observed window at time t and never future frames.
[MODE-SPECIFIC TASK] Inspect the storyboard and model state, then rank compatible candidate edits. Concept edits change the observed state or a preceding rollout state; activity-belief edits change an autoregressive source belief and affect only later forecast steps. Do not select conflicting alternatives or edits unsupported by visible evidence.
Endpoint predictions: [top-five H3 activities and probabilities]
Learned relational edges: [up to 20 kind/source/target/signed- weight tuples]
Candidate edits: [full JSON list]
Return JSON only: \{``ordered\_action\_ids'': [...], ``rationale'': ``...''\}.
\end{quote}

For Table~\ref{tab:qwen-intervention-planning}, the candidate list contains concept name, edit location (observed $t$, forecast $t{+}1$, or forecast $t{+}2$), on/off state, and current activation. 
Blind correction states that the H3 ground truth is hidden. 
In the oracle-assisted conditions, the task additionally gives the true H3 activity and, for every compatible candidate, its frozen-model probability effect, post-edit prediction, and correction status. 
One-shot planning receives effects from the unedited state and returns five edits, sequential planning receives updated conditional effects and the current selected edits, and returns one next edit.

For Table~\ref{tab:non-oracle-steering}, the task instead names the requested H3 activity and explicitly states that the future activity, future concept targets, and counterfactual effects are hidden. 
It lists all on/off concept edits and activity-belief edits across the three source steps.
The strict response schema restricts choices to listed, currently compatible candidate edits and the rationale is limited to 400 characters.

\subsubsection{Qwen intervention-policy details}
\label{app:llm-interventions}

The Qwen analysis focuses on the H3 concept-node candidate set used in Table~\ref{tab:qwen-intervention-planning}.
It contains 540 cases, comprising 30 initially correct and 30 initially wrong cases for each of three datasets and three seeds.
The blind policy receives only observed visual evidence and model state, whereas oracle-assisted policies also receive the future target and frozen-model counterfactual effects.

\paragraph{Qwen provenance and policy settings.}
Each request uses temperature $0.4$, at most 600 output tokens, and a strict JSON schema containing the ordered action IDs and rationale.
The base request seed is $20260803+100\,i+r$, where $i$ is the case index and $r$ is the policy-run index, sequential requests add $100s+a$ for budget step $s$ and retry index $a$.
The reported evaluation uses one policy run, ten concept candidates, persistent edits, and the H3 endpoint only. Malformed responses are retried under the same schema, and sequential requests add the budget step and retry index to the request seed.
The server generation defaults retain top-$k{=}20$ and top-$p{=}0.95$ because the client does not override those fields.
In the oracle-assisted condition, the policy additionally receives the held-out H3 target and the conditional single-action counterfactual effects, the blind condition receives neither.

For an explicit control, the same fixed candidate pools are evaluated with a deterministic random action order and with no edit at all.
The random order uses seed $20260805+100\,i+r$, where $i$ is the case index and $r$ is the policy-run index.
Table~\ref{tab:qwen-controls} reports the budget-five control results over the same 30 wrong and 30 correct cases per dataset and seed, the no-op row is the exact zero-edit reference at budget zero.

\begin{table}[h]
    \centering
    \caption{\textbf{Qwen intervention controls.} Mean $\pm$ sample standard deviation over three seeds at budget five. $\Delta p(y)$ is the true-class probability change in percentage points; W$\rightarrow$C and C$\rightarrow$W are cumulative correction and harm rates. The no-op row applies no intervention and is therefore exactly zero.}
    \label{tab:qwen-controls}
    \begingroup
    \scriptsize
    \setlength{\tabcolsep}{3pt}
    \begin{tabular}{lrrrrrr}
        \toprule
        & \multicolumn{3}{c}{No-op ($b{=}0$)} & \multicolumn{3}{c}{Random order ($b{=}5$)} \\
        Dataset & $\Delta p(y)$ & W$\rightarrow$C & C$\rightarrow$W & $\Delta p(y)$ & W$\rightarrow$C & C$\rightarrow$W \\
        \midrule
        BARISTA & $0.0$ & $0.0$ & $0.0$ & $\mval{-4.8}{4.4}$ & $\mval{18.9}{7.7}$ & $\mval{5.6}{9.6}$ \\
        Breakfast s1 & $0.0$ & $0.0$ & $0.0$ & $\mval{-13.7}{2.8}$ & $\mval{11.1}{3.8}$ & $\mval{46.7}{12.0}$ \\
        MPII Attr & $0.0$ & $0.0$ & $0.0$ & $\mval{-13.0}{1.4}$ & $\mval{16.7}{1.9}$ & $\mval{42.2}{1.9}$ \\
        \bottomrule
    \end{tabular}%
    \endgroup
\end{table}

Table~\ref{tab:qwen-sequential-budget-curve} reports the sequential oracle-assisted correction curve under the same Qwen policy, candidate pool, cases, seeds, and H3-only protocol described above.

\begin{table}[h]
    \centering
    \caption{\textbf{Sequential oracle-assisted Qwen correction curve.} Cumulative H3 wrong-to-correct rate in percent, mean $\pm$ sample standard deviation over three seeds. Each seed contains 30 initially wrong cases.}
    \label{tab:qwen-sequential-budget-curve}
    \begingroup
    \scriptsize
    \setlength{\tabcolsep}{3pt}
    \begin{tabular}{lrrrrr}
        \toprule
        Dataset & $b{=}1$ & $b{=}2$ & $b{=}3$ & $b{=}4$ & $b{=}5$ \\
        \midrule
        BARISTA & $\mval{23.3}{14.5}$ & $\mval{46.7}{14.5}$ & $\mval{54.4}{10.7}$ & $\mval{55.6}{9.6}$ & $\mval{57.8}{10.7}$ \\
        Breakfast s1 & $\mval{38.9}{8.4}$ & $\mval{55.6}{17.1}$ & $\mval{67.8}{10.2}$ & $\mval{72.2}{8.4}$ & $\mval{75.6}{3.8}$ \\
        MPII Attr & $\mval{41.1}{11.7}$ & $\mval{51.1}{12.6}$ & $\mval{55.6}{13.5}$ & $\mval{56.7}{11.5}$ & $\mval{58.9}{7.7}$ \\
        \bottomrule
    \end{tabular}%
    \endgroup
\end{table}

The sequential correction rate increases monotonically across all three datasets.
At budget five, sequential Qwen shares approximately 3.2 of its five selected actions with the greedy sequential oracle on average, compared with approximately 2.1 for one-shot Qwen.
Twenty-five of 2,700 Qwen decisions use the logged conditional-best fallback after Qwen repeatedly proposes unavailable edits, so this setting is an oracle-assisted robustness analysis rather than a deployable intervention policy.

\section{Concept grounding of forecast states}
\label{app:concept-grounding}

Gemma-3-27B (a different model family than the vocabulary generator) judged concept presence in the real observed, H1, and H3 frames of 300 windows (4,500 judgments), and one author labelled a blinded subset of 240 (with a single annotator, the human numbers are only indicative). Table~\ref{tab:grounding} shows that forecast states are as grounded as the observed scores they are trained on. After edits, 99.5--100\% of unedited concepts keep their label.

\begin{table}[h]
    \centering
    \caption{\textbf{AUROC of concept states.} Gemma rows: BARISTA / Breakfast s1 / MPII; human row pooled (TRACE H1+H3: 0.74, 95\% CI 0.64--0.84).}
    \label{tab:grounding}
    \scriptsize
    \begin{tabular}{lccc}
        \toprule
        Labels & Observed & TRACE H1 & TRACE H3 \\
        \midrule
        Gemma judge & 0.53 / 0.64 / 0.58 & 0.61 / 0.61 / 0.58 & 0.54 / 0.65 / 0.59 \\
        Human (Gemma itself: 0.69) & 0.75 & 0.71 & 0.77 \\
        \bottomrule
    \end{tabular}
\end{table}

\end{document}